\documentclass[journal, 10pt]{IEEEtran}
\usepackage{graphicx}
\usepackage{amssymb}
\usepackage{amsmath}
\usepackage{cite}
\usepackage{amsthm}
\usepackage{epstopdf}
\usepackage{color}
\usepackage{soul}
\usepackage{balance}
\usepackage[caption=false]{subfig}
\usepackage{multirow}
\usepackage{multicol}
\usepackage{booktabs}
\usepackage{amsfonts}
\usepackage{epsfig}
\usepackage{url}
\usepackage{algorithm}
\usepackage{mathrsfs}
\usepackage{amsmath}
\usepackage{physics}
\usepackage{hyperref}

\usepackage[dvipsnames]{xcolor}
\definecolor{duongfix}{rgb}{0,0,0}
\definecolor{firstreview}{rgb}{0,0,0}

\usepackage[english]{babel}
\usepackage[utf8]{inputenc}
\usepackage{textcomp}
\usepackage{algorithm}
\usepackage[noend]{algpseudocode}

\newtheorem{theorem}{Theorem}

\makeatletter
\def\ScaleIfNeeded{%
\ifdim\Gin@nat@width>\linewidth \linewidth \else \Gin@nat@width \fi
} \makeatother

\definecolor{fix}{rgb}{0,0,0}  

\begin{document}



\title{HCFL: A High Compression Approach for Communication-Efficient Federated Learning in Very Large Scale IoT Networks}  


\author{Minh-Duong~Nguyen,
        Sang-Min~Lee,
        Quoc-Viet~Pham,~\IEEEmembership{Member,~IEEE}, 
        Dinh~Thai~Hoang,~\IEEEmembership{Member,~IEEE},
        Diep~N.~Nguyen,~\IEEEmembership{Senior Member,~IEEE},
        and Won-Joo~Hwang,~\IEEEmembership{Senior Member,~IEEE}
        
\thanks{Minh-Duong Nguyen and Sang-Min Lee are with the Department of Information Convergence Engineering, Pusan National University, Busan 46241, Republic of Korea (email: \{duongnm@pusan.ac.kr, pms88520@pusan.ac.kr\}).}
\thanks{Quoc-Viet Pham is with the Korean Southeast Center for the 4th Industrial Revolution Leader Education, Pusan National University, Busan 46241, Republic of Korea (e-mail: vietpq@pusan.ac.kr).}
\thanks{Dinh~Thai~Hoang, and Diep~N.~Nguyen are with the School of Electrical and Data Engineering, University of Technology Sydney, Sydney, NSW 2007, Australia (e-mail: \{hoang.dinh, diep.nguyen\}@uts.edu.au).}
\thanks{Won-Joo Hwang is with the Department of Biomedical Convergence Engineering, Pusan National University, Yangsan 50612, Republic of Korea (e-mail: wjhwang@pusan.ac.kr).}
\thanks{This work was supported by a National Research Foundation of Korea (NRF) Grant funded by the Korean Government (MSIT) under Grants NRF-2019R1C1C1006143 and NRF-2019R1I1A3A01060518.}
}
\maketitle

\begin{abstract}
Federated learning (FL) is a new artificial intelligence concept that enables Internet-of-Things (IoT) devices to learn a collaborative model without sending the raw data to centralized nodes for processing. Despite numerous advantages, low computing resources at IoT devices and high communication costs for exchanging model parameters make applications of FL in massive IoT networks very limited. In this work, we develop a novel compression scheme for FL, called \textit{high-compression federated learning (HCFL)}, for very large scale IoT networks. HCFL can reduce the data load for FL processes without changing their structure and hyperparameters. In this way, we not only can significantly reduce communication costs, but also make intensive learning processes more adaptable on low-computing resource IoT devices. Furthermore, we investigate a relationship between the number of IoT devices and the convergence level of the FL model and thereby better assess the quality of the FL process. We demonstrate our HCFL scheme in both simulations and mathematical analyses. Our proposed theoretical research can be used as a minimum level of satisfaction, proving that the FL process can achieve good performance when a determined configuration is met. Therefore, we show that HCFL is applicable in any FL-integrated networks with numerous IoT devices.
\end{abstract}

\maketitle

\section{Introduction}
\label{section:2}

The rapid increase in Internet-of-Things (IoT) applications has revolutionized the big data and machine learning (ML) technology fields \cite{hussain2020machine}. Conventionally, ML and big data algorithms are deployed in the centralized servers \cite{Goodfellow-et-al-2016}. This central deployment can be considered as an effective approach if all the data are available in one consolidated database. Therefore, the whole big data distribution can be processed without any problems such as non independent and identically distributed (non-IID) and lack of sampling population, which is unable to represent the whole data distribution. The arrival of the golden era of massive IoT, in addition to the development of the state-of-the-art 5G and future 6G wireless systems, has expedited the demands for distributed learning solutions at the edge of the network \cite{chen2021massive, de2021survey}. 

FL \cite{mcmahan2017communicationefficient}, an emerging AI paradigm, has recently attracted considerable attention from the research communities thanks to its nice features and wide-ranging applications. In conventional ML techniques, a server collects the user data and performs a computing process centrally, thereby increasing the risk of high communication overload and data leakage. FL, on the other hand, aims to mitigate the privacy concerns by locally training the ML models, and thus the users only need to dispatch to the server their model parameters \cite{nguyen2021federated}. Therefore, this paradigm preserves the private and collaborative aspects of the ML framework for distributed users. As a result, a number of AI applications leveraging the FL process for IoT networks have been developed recently, such as vehicular cloud \cite{10.1145/3372224.3418168}, remote applications \cite{10.1145/3372224.3419186}, smart healthcare \cite{chen2020fedhealth}, and mobile edge networks \cite{lim2020federated}.

Despite recent advancements in computing hardware and computing paradigms such as fog computing and
mobile edge computing (MEC), the limitation of communication resources still remains as an obstacle in IoT systems \cite{pham2019coalitional}. With the fact that the number of IoT devices is increasing dramatically in the modern world and the wireless resources are limited, communication proficiency sets off to be one of the key challenges for carrying out massive IoT scenarios in which the large-scale FL system is integrated. 
Therefore, some approaches to address the communication efficiency problem for FL processes have been introduced recently \cite{chen2021communication}. For example, the authors in \cite{10.1145/3372224.3419188} utilize the submodel framework, wherein the clients decide which parts are necessary for the model update process instead of implementing the whole model. The FedBoost algorithm in \cite{pmlr-v119-hamer20a} is a combination of an ensemble learning and the random sampling theory. To be more specific, FedBoost selects a random partial set of clients to deliver the updates instead of sending all parameters to every client in each round. The secure aggregation approach proposed in \cite{9049066} and \cite{10.1145/3133956.3133982} utilizes the random sampling method to cross-validate the model updates between different pairs of clients before forwarding the model parameters to the aggregation module at the server. Furthermore, the authors in \cite{9164912} and \cite{9145588} formulate the optimization problems to minimize the time cost for each communication round \cite{9164912} and maximize the aggregation data rate from FL users \cite{9145588}. \textcolor{firstreview}{In order to reduce the communication load during the propagation phase, many researches have applied sparsification and model pruning into FL. Authors in \cite{2020-FL-prunnedFL} inherit the simple regularization method to develop pruneFL algorithm. This algorithm can reduce the number of  parameters up to $7$ times. PruneFL, on the other hand, requires more than $6000$ communication rounds to reach convergence, which is inappropriate for IoT networks. CA-DSDG \cite{2020-FL-CADSDG} and CE-FedAvg \cite{2020-FL-CEFedAvg} both employ the same sparsification concept, in which model parameters that do not change much are dropped from the updating phases. However, the efficiency by sparsification is not convincing where the achieved compression ratio is capped at $70\%$. Another effective method which is both efficient for communication and computation is model quantization. The works in \cite{xu2020ternary, Li2016TernaryWN, rastegari2016xnornet} focus on model compression wherein the entire network structure is estimated and incorporated into a new model with a low definition model parameter set. The full-precision weights are estimated into approximate set with ternary format (\textcolor{duongfix}{which includes only three values: $-1$, $0$, $1$}) \cite{DBLP:journals/corr/ZhuHMD16}. Since the deep network parameters are quantized into 2-bit values, the method already reaches its compression limitation. Therefore, the maximum compression efficiency that this method can achieve is constrained by $90\%$.}

In order to further improve the communication efficiency of FL processes, in this work, we propose a new compression strategy, called \textit{high-compression federated learning (HCFL)}, for IoT networks with a very large number of IoT devices. Remarkably, thanks to the compact design on the presentation layer of the OSI model \cite{2012-ComputerNetworking}, HCFL is strictly in compliance with the entity-specific information flow between current 3GPP interfaces while being communication efficient. Thus, our proposed HCFL scheme can be cooperated with other communication-efficient methods (e.g., quantization \cite{Li2016TernaryWN, xu2020ternary, 2018-DL-Autoencoder-Quantization}, sparsification \cite{2020-FL-CADSDG}, model pruning \cite{2020-FL-prunnedFL, 2020-FL-CEFedAvg}) to improve the performance of the whole FL process. We can sum up our contributions in this paper as follows:
   \begin{itemize}
     \item We introduce a novel compression scheme, called \textit{HCFL}, for FL. We leverage an under-complete autoencoder structure in the FL process to develop a communication-efficient strategy that can reduce the transmission load of the FL mega-system with a very large number of IoT devices.
     
     \item We provide theoretical analysis of the proposed HCFL method under the FL mega-system. We demonstrate that the HCFL method can achieve a very high performance, e.g., the models with HCFL can be compressed more than $95\%$ of their original size. \textcolor{firstreview}{We also prove that the lossy compression by HCFL adds noise into the FL training process, which encourages exploration efforts for our large-scale {IoT} learning system. Thus, HCFL can improve the FL's  overall performance and reduce FL's both convergence round and delays.} Furthermore, we investigate a relationship between the system performance and the other compress-aided FL process's characteristics such as the number of IoT devices and the compressor reconstruction error. 
     
     \item We conduct various simulations with an in-depth comparison with the conventional \textcolor{duongfix}{FL process} in terms of performance and communication efficiency. We evaluate the proposed HCFL method under different conditional simulations, including compression efficiency at different compression ratios, convergence analysis at distinct FL process and client predictor settings. The simulation results reveal that the HCFL method can achieve a compression ratio up to $32$ times better than that of the conventional FL process. Notably, the trade-off between compression ratio and performance is acceptable when the accuracy loss on the HCFL system is less than $3\%$ in comparison with the baseline method.
   \end{itemize} 
   
The rest of this article is organized as follows. Section~\ref{sec:background} briefly reviews the fundamental knowledge about FL and autoencoder concept. Section~\ref{sec:hcfl} demonstrates the theoretical analysis of HCFL system and how the HCFL and FL process support each other. Section~\ref{sec:algorithm} details the deployment strategy of the compression-aided federated system along with two appropriate compression model structures. Simulations conducted to compare the performances of a standard FL process and the proposed method in terms of global loss, data reconstruction error and convergence rate are described in Section~\ref{sec:simulation}. Finally, conclusions are presented in Section~\ref{sec:conclusion}.
%

\section{Background and Methods} 
\label{sec:background}
This section first presents the system model, then provides some preliminaries of the conventional FL procedure and its essential formulations, and finally shows the main features and definitions of the autoencoder model.

\subsection{System Model}\label{SM}

\begin{figure}[t]
\centering
\includegraphics[width=0.975\linewidth]{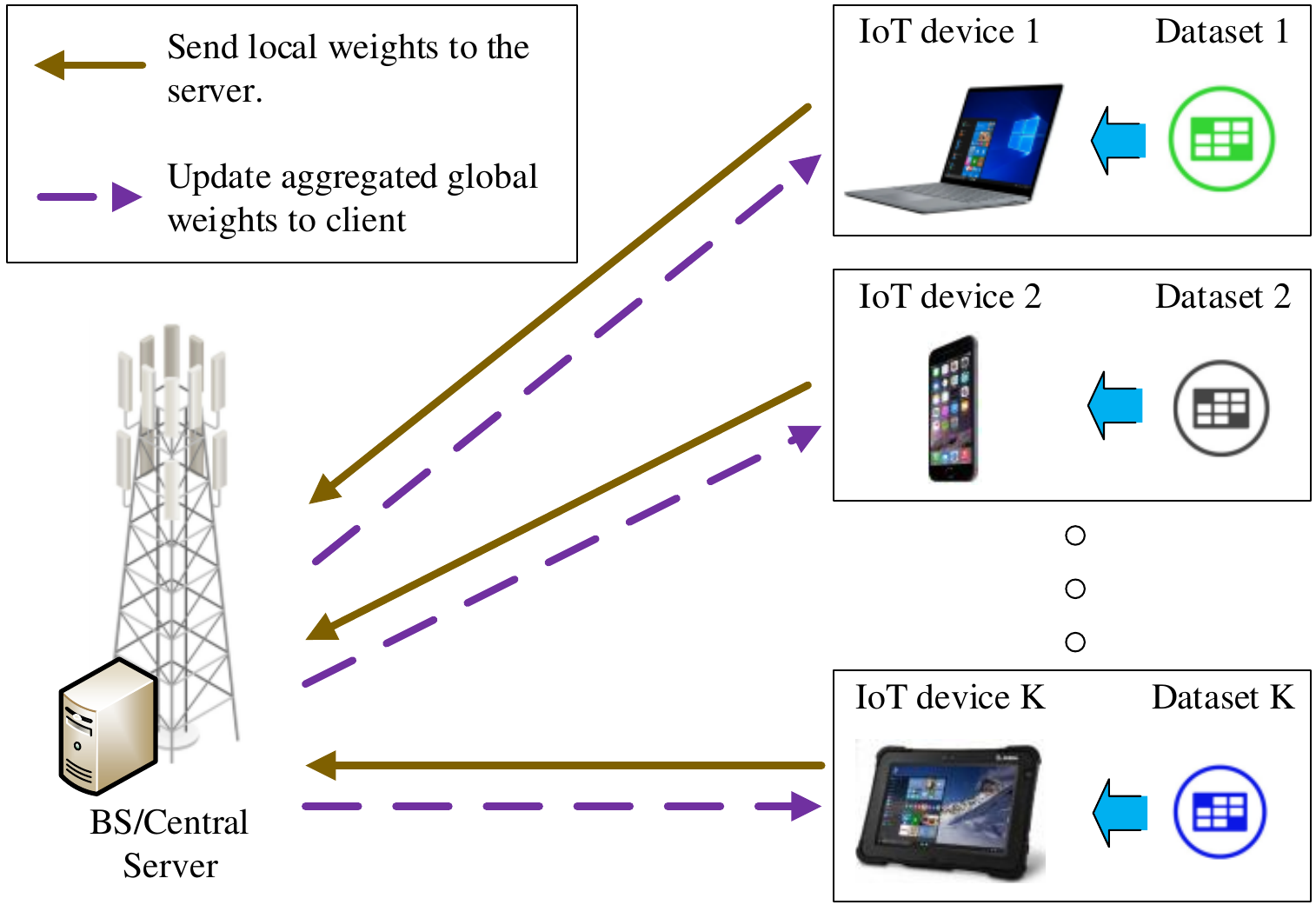}
\caption{\textcolor{duongfix}{An illustration of FL. The users train the local predictors on their distributed devices, and then the trained local neural networks are sent to the server. The server then executes the aggregation to generate the global model parameters. Lastly, the latest global model is uploaded back to the users, and the process continues until the global model converges.}}
\label{federated-learning-architecture}
\end{figure}

We consider a mega FL-integrated IoT system involving one server and a set 
of $K$ clients (i.e., IoT devices), as illustrated in Fig.~\ref{federated-learning-architecture}. Clients and server jointly build a shared ML model for processing and data analysis. Each client collects data from its sensors or users. We assume that all the clients' datasets have the IID property in this study \cite{mcmahan2017communicationefficient}. The client's data is processed through an ML-based predictor on each device distributively. Because a large number of clients participate in each communication round, every participating client needs to share limited bandwidth resources. Thus, the transmission rate of each client is also limited. This problem leads to the need for communication efficiency in FL-integrated IoT networks and motivates us to complete this work. Our work aims to reduce the communication burden significantly in the meta FL-integrated IoT system.

\subsection{Federated Learning}\label{AA}
FL is an up-to-the-minute concept of ML. In FL, instead of collecting the data and utilizing them for central training at one location, the data is processed distributedly at the clients. 
The cooperation between the server and clients is to ensure that the model losses are verified at the server, and the improvement is updated onto the distributed clients gradually, as shown in Fig.~\ref{federated-learning-architecture}.
\textcolor{duongfix}{The FederatedAveraging (FedAvg) algorithm, proposed in \cite{mcmahan2017communicationefficient}, utilizes} a sum averaging function to combine loss and weight values from different clients as follows:
\begin{equation}
\begin{split}
f(w) = \frac{n_k}{n}\sum^{K}_{k=1}{F_k(w)}  
\quad
\textrm{where} 
\quad
F_k(w) = \frac{1}{n_k}\sum_{i\in{P_k}}f_i(w),
\end{split} 
\end{equation}
Here, $w$ denotes the parameters of the FL model, 
\(F_k(w)\) and \(f_i(w)\) denote the loss value on the client \(k\) and each data point of the whole dataset with a size $n$, respectively. The dataset is distributed over \(K\) clients with a data size of \(n_k\) for each client in FL. The gradient on the client $k$ is given as \(g_k = \delta{F_k(w_k)}\). \textcolor{duongfix}{The aggregation function at the server generates a close-form function with the loss sum-averaging update as follows:}
\begin{equation}\label{PMFA1}
\begin{gathered} 
w_{t+1} \leftarrow \sum^{K}_{k=1}{\frac{n_k}{n}w^k_{t+1}}. 
\end{gathered}
\end{equation}
The equality \(\frac{n}{n_k}=K\) is valid when the dataset on each client is IID with each other on the whole set of dataset of the system \cite{9187874} and every client samples the same amount of data each communication round. Thus, the FedAvg algorithm updates the model weights as follows:  
\begin{equation}\label{PMFA2}
\begin{gathered} 
w_{t+1} \leftarrow \frac{1}{K}\sum^{K}_{k=1}{w^k_{t+1}}.
\end{gathered}
\end{equation}
\subsection{Autoencoder}\label{BB}
Autoencoders utilized in deep learning \cite{Goodfellow-et-al-2016} are dual-symmetric neural networks designed to produce the output data that is approximately-but-not-equal to the intake of that system. Autoencoders learn a concentrated representation of the input while the most valuable information is retained.
There are two components in the aforementioned system: an encoder that processes a transformation: $h=f_\omega(x),$ and a decoder $\hat{x}=g_{\omega'}(h)$ that is responsible for the reconstruction.
The \(f\) and \(g\) represent the encoder and decoder neural networks, respectively, and \(h\) denotes the feature vector of the model in the aforementioned functions. The encoder \(f\) transforms the incoming matrix into a completely divergent model, whereas the decoder \(g\) is tasked with rehabilitation. The parameters of the proposed network is learned simultaneously on the task of reconstructing the original output. 

The loss function \(\mathcal{L}(w)\), which is a measure of the discrepancy between the input \(x\) and the output \(\hat{x}\), is applied to obtain the lowest possible reconstruction error \cite{bengio2014representation}, which is defined as follows: 
\begin{equation}\label{LF1}
\begin{split}
\mathcal{L}(w) &= \frac{1}{2}\norm{\hat{x} - x}^2_2 = \frac{1}{2}\sum^{N}_{k=1}{(\hat{x} - x)^2}.
\end{split}
\end{equation}
The proposed back-propagation operation uses the conventional mean squared error (MSE) loss function \cite{Goodfellow-et-al-2016} to compute the Euclidean distance between the two vectors of the output and input of the autoencoder. 

\begin{figure}[t]
\centering
\captionsetup{justification=centering}
\includegraphics[width = 0.95\linewidth]{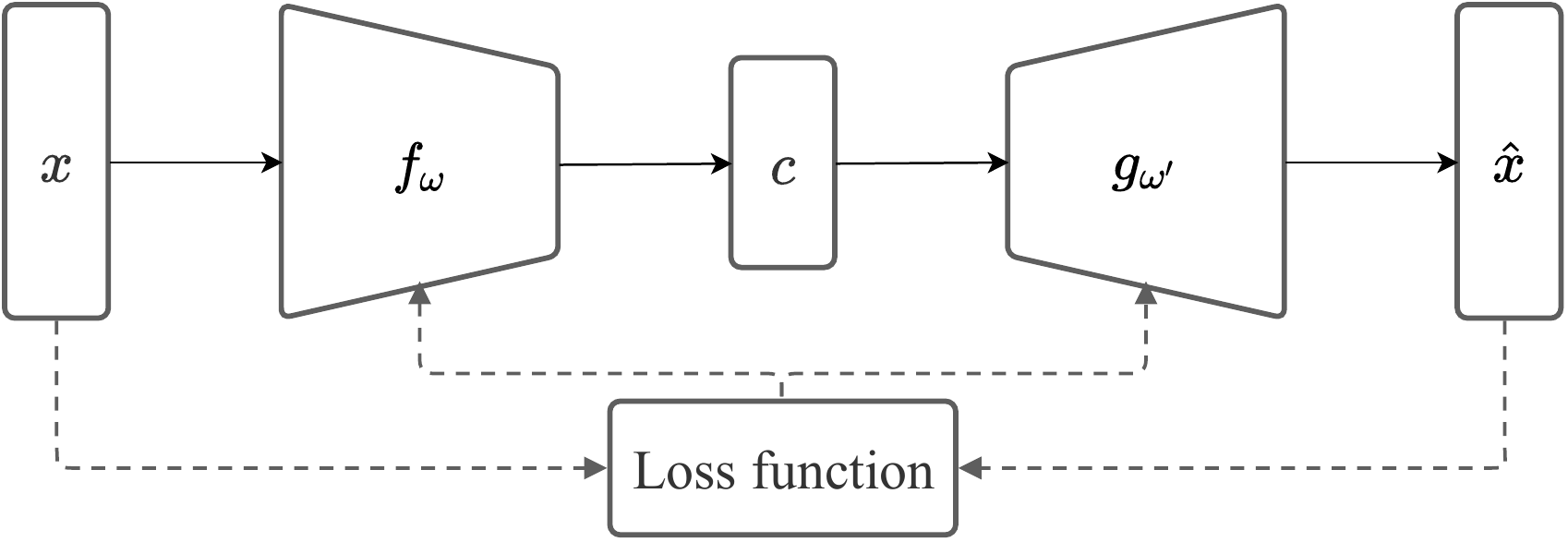}
\caption{Structure of a general autoencoder.}
\label{ae}
\end{figure}

The illustration of a general autoencoder is shown in Fig.~\ref{ae}. Autoencoders are designed to not copy the output from the input perfectly, as aforementioned. Usually, restrictions are conventionally imposed to make the distribution of the output data to be exactly the same with that of the original information. Because of the objective to prioritize the input aspects, which can represent the whole data distribution, the encoder manages to collect the valuable properties only.
The main objectives of autoencoders are conventionally feature extraction, dimensionality reduction, and data augmentation. Autoencoders learn the data characterization in a reduced dimensional space by \textcolor{duongfix}{placing additional focus on essential features} while attempting to eliminate redundancy and noise. It is based on the encoder-decoder architecture, wherein the encoder encrypts the data with a high dimensional space to the lower-dimensional version. Moreover, the decoder performs in a reverse way, i.e., the decoder reconstructs the initial high-dimensional information from the reduced-dimensional data.

\section{Proposed Compression-Aided FL Algorithm} 
\label{sec:algorithm}
\textcolor{firstreview}{In this section, we present how to implement our proposed HCFL-integrated FL in massive IoT networks. We first propose the problem formulation for our HCFL method. Then, we propose the system deployment of HCFL in an FL process and clarify the procedure of the HCFL-aided FL.} Secondly, we find that when implementing a traditional multi-layered autoencoder, the performance of the HCFL model is degraded due to the curse of dimensionality. Therefore, we propose new model architectures for the compressor and extractor that improve the integration efficiency of the HCFL into the FL process. Furthermore, we reveal how to pre-process the input model weights and how the model is trained given the model parameter dataset.

\begin{figure}[t]
\centerline{\includegraphics[width=0.75\linewidth]{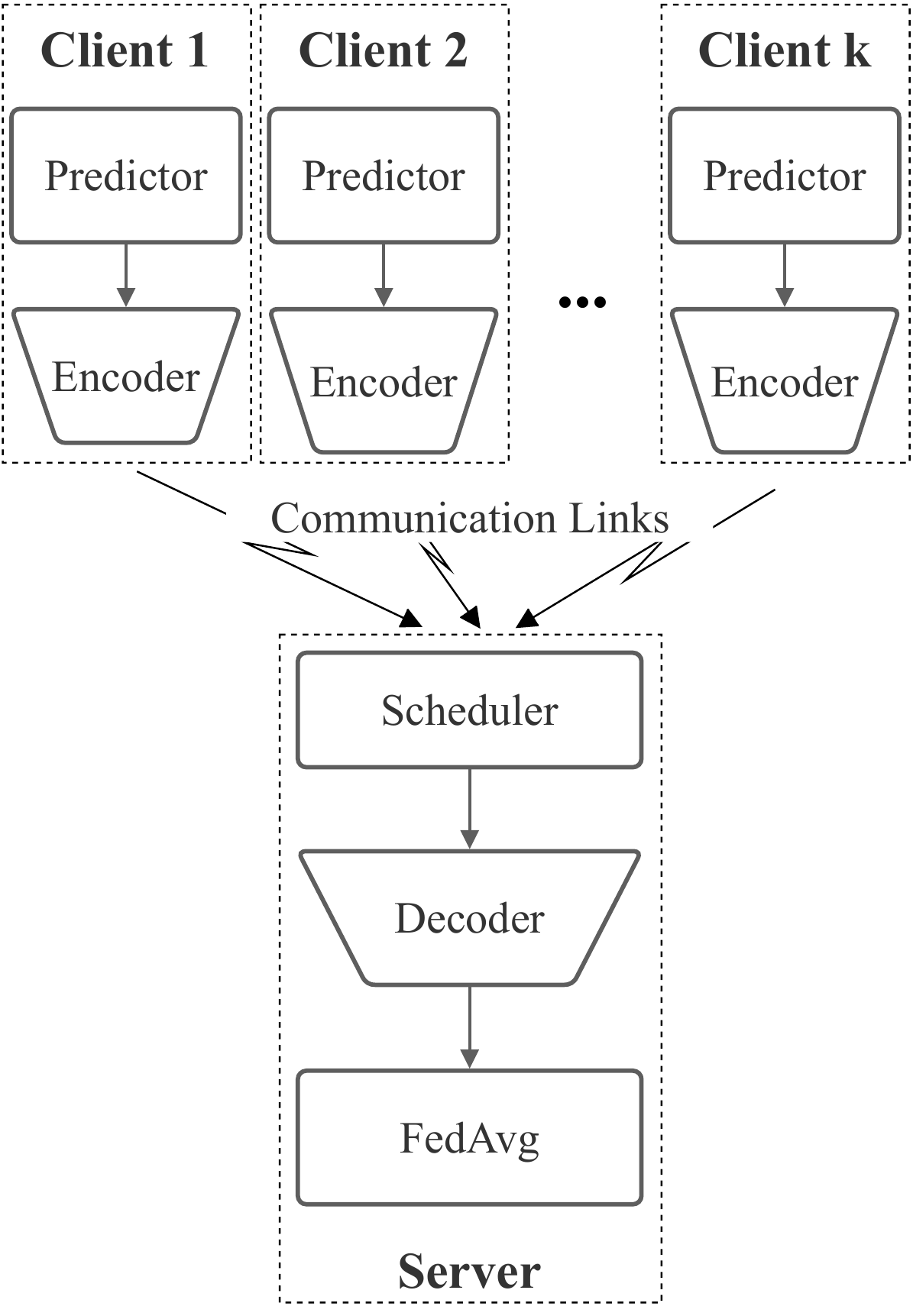}}
\caption{HCFL system deployment.}
\label{sys-deploy-1}
\end{figure}

\subsection{Problem Formulation}
\label{subsec:problem-formulation}
In this part, we aim to develop the framework to minimize the reconstruction error between the original model at input and reconstructed model at output for HCFL in the FL process. In general, HCFL operates as an autoencoder to compress the distributed clients' model and reconstruct the encoded model. Nevertheless, because the clients apply gradient descent stochastically, the models' weight distribution transformation remains undercover. To acknowledge those problems, we need to capture the generalization and transforming trends of every model's distributions, especially when the model's gradient descent is operated in different directions. Thus, our proposed HCFL should be satisfied the two following conditions: 
\begin{itemize}
    \item Minimize the reconstruction error between the original model and reconstructed model as mentioned in \eqref{LF1}.
    \item Maximize the mutual information between the original model and encoded data. Therefore, we can maximize the information transferred from the original data to the encoded data, which improves the performance of the encoder and thus enhance the decoder quality. The optimization problem problem is introduced as follows: 
    \begin{equation}\label{eq:HCFL-encoded-info-maximize}
    \begin{gathered}
        \max_{\theta}{I(W,C) = D_\text{KL}\big[p(W,C) \parallel p(W) \otimes p(C)\big]},
    \end{gathered}
    \end{equation}
    where $\otimes$ is the Kronecker product between two matrices $W$ and $C$, $\theta$ is the model parameters of the HCFL model, and the $D_\text{KL}$ stands for relative entropy between two data distribution. 
\end{itemize}
In this way, we can formulate a joint optimization problem to solve both of the aforementioned tasks. Meanwhile, to conjugate the distance-based problem in \eqref{LF1} and entropic problem in \eqref{eq:HCFL-encoded-info-maximize}, we consider the relationship between MSE and cross entropy (CE). We follow an assumption that the output of deep network in HCFL is demonstrated as Gaussian distribution with variance of $\sigma^2$ \cite{Goodfellow-et-al-2016}: 
\begin{equation}\label{eq:HCFL-output}
\begin{split}
    p(\hat{W} \mid W) &= \mathcal{N}(W, \sigma^2) \\ 
                      &= \frac{1}{\sqrt{2\pi \sigma^2}}\text{exp}\left( -\frac{1}{2} \frac{(W-\hat{W)^2}}{\sigma^2} \right).
\end{split}
\end{equation}
Applying CE on the HCFL's output, we have: 
\begin{equation}\label{eq:HCFL-CE}
\begin{split}
   & H(W,\hat{W}) = -E_{p(W)} \log p(\hat{W}) \\ 
   &= -E_{p(W)}\log\Big[ \frac{1}{\sqrt{2\pi \sigma^2}}\text{exp}\Big( -\frac{1}{2} \frac{(W-\hat{W)^2}}{\sigma^2} \Big) \Big] \\
   &= \mathcal{O} \big(-E_{p(W)}(W-\hat{W})^2\big).                
\end{split}
\end{equation}
$\mathcal{O}$ describes the growth rate of the function, $E_{p(W)}$ is the expectation over distribution of model parameters $W$. From the above formula, we can see that $H(W, \hat{W})$ is proportional with MSE loss $E_{p(W)}(W-\hat{W})^2$. Thus, we have the joint problem formulation as follows: 
\begin{subequations}
	\label{eq:HCFL-joint-problem}
	\begin{alignat} {3}
		& \min_{\theta} ~~~
		&	 &  L = H - I
		\label{subeqn-HCFL:opt-pro}       \\
		& \text{s.t.}
		&	& H = \lambda H(W,\hat{W}),
		\label{subeqn-HCFL:CE}\\ 
		&   &   & I = (1-\lambda) D_\text{KL}\big[p(W,C) \parallel p(W) \otimes p(C)\big],
		\label{subeqn-HCFL:RE} 
	\end{alignat}
\end{subequations}
\textcolor{fix}{where $\theta$ is the parameter set of the HCFL model and $\lambda$ is the scale coefficient between two tasks $H$ and $I$. Intuitively, the problem in \eqref{eq:HCFL-joint-problem} is the alternative version of bottleneck technique. Therefore, the choice of $\lambda$ is similar to the scaling factor choice in \cite{2000-DL-InformationBottleNeck, 2016-DL-DeepVariation-InformationBottleNeck}}. The HCFL is thus trained by the following gradient descent process: 
    \begin{equation}\label{eq:HCFL-update}
    \begin{split}
        \theta = \theta - \gamma \nabla L,
    \end{split}
    \end{equation}
where $\gamma$ is the learning rate for the HCFL training process.

\subsection{System Deployment}\label{X0}
This part discusses the system deployment of the HCFL in the FL. As we can see from Fig.~\ref{sys-deploy-1}, the HCFL comprises two components: encoders which are located on each client's domain, and a single decoder which is embedded in the server's firmware. Although numerous encoders are required for the activation of the compression operation on the clients, the server's side only requires a single decoder because the incoming client-training-information from all clients is discontinuous and can be scheduled using the First-In-First-Out rule. This ensures that the hardware requirement of the server can still be satisfied. 

Algorithm~\ref{HCFL} shows the complete pseudo-code for the HCFL-integrated FL. After the initialization, the server starts its training iterations. Each iteration is named as one communication round. At every communication round, the server uploads global weights to every client and calls for updates from $K$ selected clients. The selected clients train their models in parallel and then send their new local models to the server. The server decodes all of the received local parameters and then executes the accumulation and averaging to acquire a new global model. This process is continuously processed until the FL achieves the desired convergence or when the max communication round value is reached. \textcolor{firstreview}{Moreover, the encoder compresses the model prior to the packaging and modulating phases of the communication process on the transmitting links. Likewise, the decoder reconstructs the model after the demodulation and de-packaging phases in the receiving links. This closed-loop process ensures the compatibility of HCFL in any IoT systems.}


\subsection{HCFL Network Architecture}\label{X2}
The proposed HCFL framework is expected to compress the model parameter data with a ratio of $1$:$4$ to $1$:$32$ in terms of size. \textcolor{duongfix}{This rate can be achieved by dealing} with the loss of data during the encoding scheme. Hence, a deep neural network is leveraged to reduce the loss between the reconstructed and the original data. 

\subsubsection{Definition of Datasets} To develop HCFL, we first define the dataset for the network. The data are extracted from the model parameter dataset. Instead of extracting the model parameters towards the end of the training, the data prepared for this system is generated after each epoch in each client to assist the compressor in learning the values and spatial distributions of the neural network‘s coefficients. The curse of dimensionality and the computation cost at the client's end were reduced by splitting the data into two components to be trained with two different compressors, namely the convolution kernel dataset and dense network dataset, \textcolor{duongfix}{where elements in each part have the similar dispensational characteristics.}

\begin{algorithm} [t]
\caption{HCFL compression-aided FedAvg FL. We denote $k$ to be the client index in a total number of $K$ clients; $E$ is the local epochs, $\eta$ is the learning rate, and the local mini-batch size is represented as $B$.} 
\label{HCFL}
\begin{algorithmic}
\Procedure{ServerExecutes}{}
\State Initialize \(w_0\)
\For{each round \(t=1,2,\dots\)}
    \State $m\gets \max(1,K \times C)$
    \State $S_t\gets$ (random set of $m$ clients) 
    \For{all $k\in S_t$ \textbf{in parallel}}
        \State $h^k_{t+1} \gets$ \textsc{ClientUpdates}($w_t,k$)
        \State $w^k_{t+1} \gets$  \textsc{Decode}($h^k_{t+1}$)
        \State $w_{t+1} \gets \frac{(k-1)}{k}w_{t+1}+\frac{(1)}{k}w^k_{t+1}$
    \EndFor
\EndFor
\State Updates $w$ to the clients.
\EndProcedure
\end{algorithmic}
\begin{algorithmic}
\Procedure{ClientUpdates}{$w,k$} 
\State $B\gets$ (divide $P_k$ into batches of size B)
\For{each local epoch $e$ from 1 to $E$}
    \For{batch $b \in B$}
        \State $w \gets w - \eta\triangledown l(w;b)$
    \EndFor
\EndFor
$h$ = Encode($w$)
return $h$ to server
\EndProcedure
\end{algorithmic}
\end{algorithm}

\begin{figure*}[t]
	\centering
	\captionsetup{justification=centering}
	\subfloat[Architecture of HCFL compressor.\label{fig:HCFL-compressor}]{\includegraphics[width=0.435\linewidth]{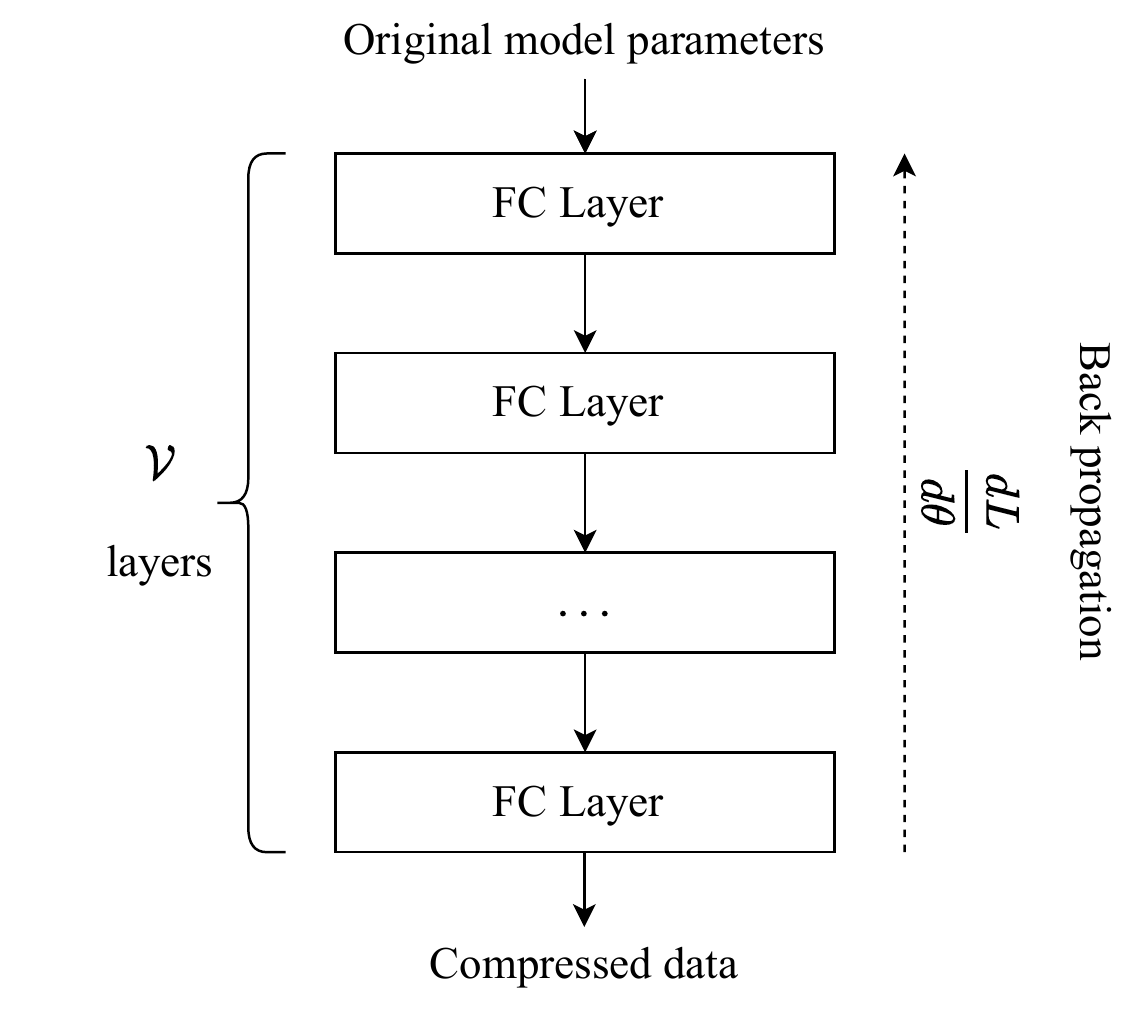}}\quad
	\subfloat[Architecture of HCFL extractor.\label{fig:HCFL-extractor}]{\includegraphics[width=0.435\linewidth]{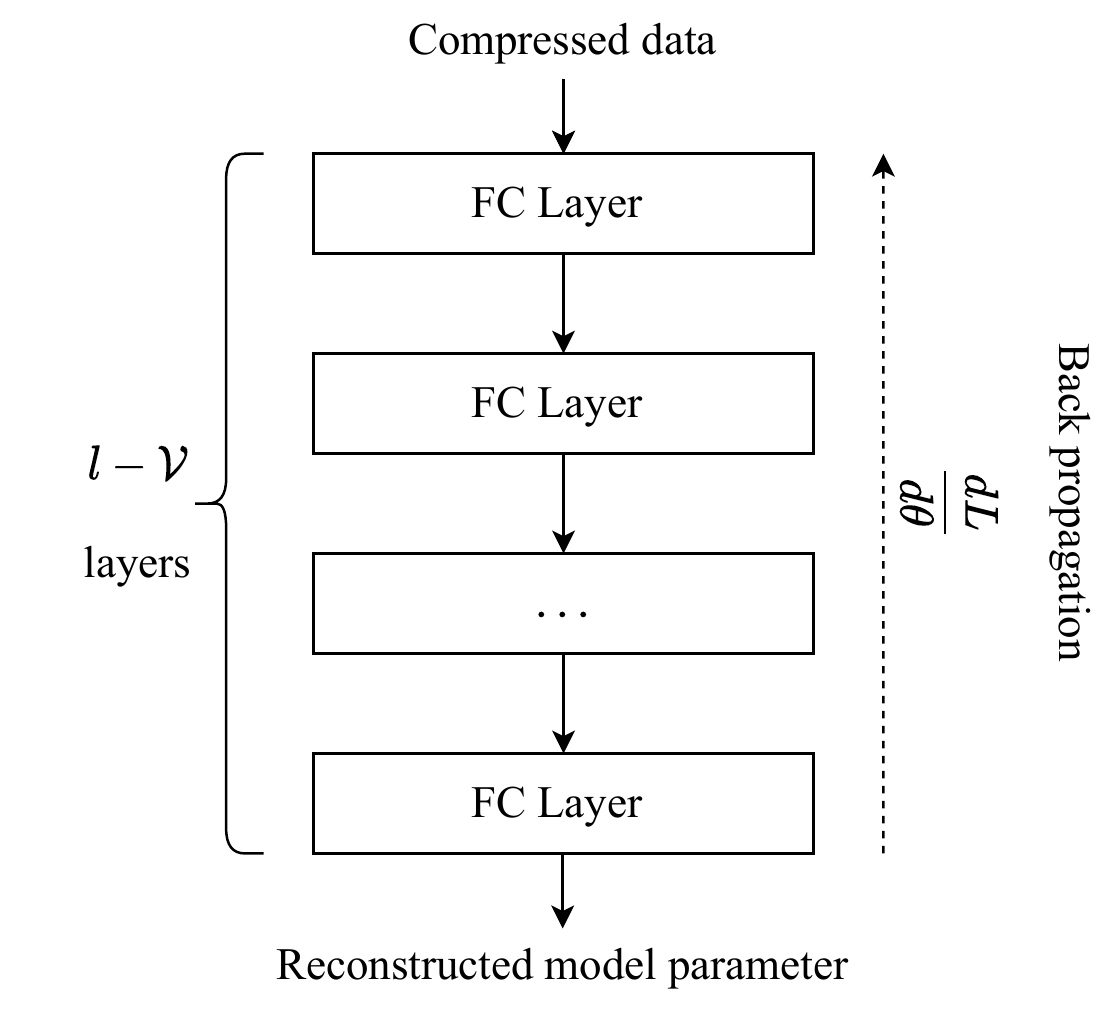}} 
	\caption{HCFL network architecture. There are $\mathcal{V}$ FC layers on the compressor and $l-\mathcal{V}$ FC layers on the extractor, respectively. The FC layers in each HCFL part is connected sequentially.}
	\label{Fig:Perf_AAS}
\end{figure*}

\subsubsection{The Proposed Compression System} 
\begin{figure}[t]
\centerline{\includegraphics[width=0.5\linewidth]{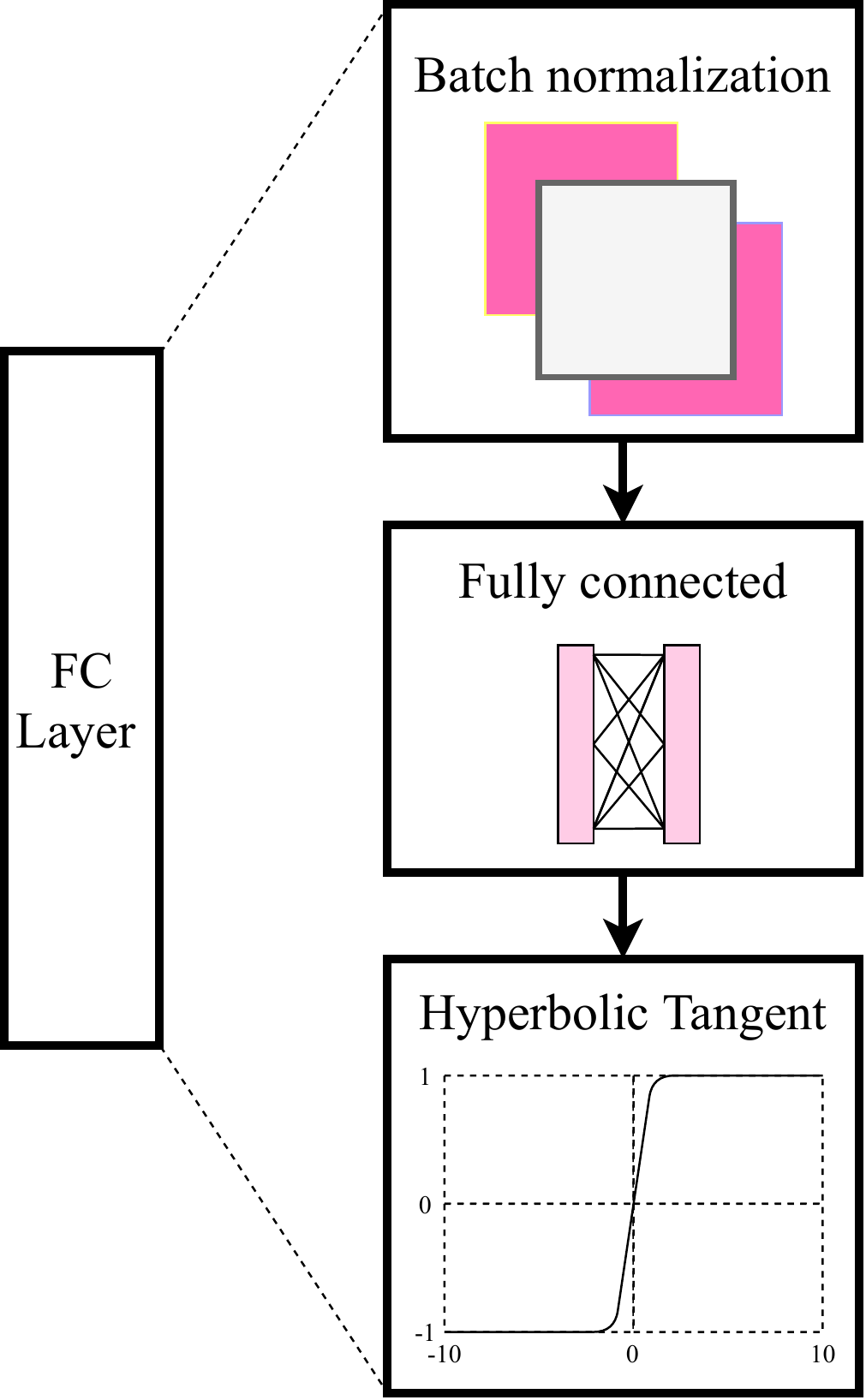}}
\caption{Composition of an FC layer.}
\label{fig:FC-layer}
\end{figure}

uses $\mathcal{V}$ \textcolor{duongfix}{fully-connected} (FC) layers at the encryptor (Fig.~\ref{fig:HCFL-compressor}) and $(l-\mathcal{V})$ FC layers at the extractor (Fig.~\ref{fig:HCFL-extractor}), where $l$ is the total hidden layers of the HCFL system to activate the dimensionality reduction. Each FC layer consists of a dense layer followed by a Tanh activation function for each layer node. The usage of Tanh is to guarantee the output of the HCFL in the range $[-1,1]$\textcolor{duongfix}{, which is the value range of original model parameters}. The FC layer also uses an additional batch normalization in the input in order to make the HCFL more stable and faster by re-centering and re-scaling, as in Fig.~\ref{fig:FC-layer}. The depth of the HCFL hidden layer is set according to the compression ratio of the HCFL and the complexity of the input model. The higher the HCFL compression ratio is, the more FC layers are added into the deep network. As more layers being added to the neural network, the lower bound of the log probability that the model assigns for the training data ascends (which will be presented in Section~\ref{sec:HCFL-settings}). Thus, the deep neural networks deliver better compression performance than those of shallow or linear neural networks \cite{Hinton2006ReducingTD,10.1162/neco.2006.18.7.1527}.

\subsubsection{Data Pre-processing} A detailed description is given below: 
\begin{itemize}
    \item \textbf{Data preparation:} The model parameters are stored during the FL execution. To avoid the dataset imbalance, we only fetch the pre-saturated client's predicting models. Moreover, with the proposed method, the HCFL compressor can learn the model general distribution at every learning state.   
    \item \textbf{Data segmentation:} We apply the divide-and-conquer algorithm \cite{10.5555/1614191} to break down each individual model parameter into two or more sub-datasets whose distributions have the high mutual information. Therefore, the clustered dataset distributions become simple enough that HCFL can avoid the curse of dimensionality. The performance analysis of the data segmentation technique is proposed in Section~\ref{theorem:FL-model-complexity}.
\end{itemize}
\subsection{Proposed Training Phase} We adopt the transfer learning technique to implement the HCFL. Firstly, we train a pre-model with a small amount of dataset on the server. By applying augmentation on small dataset on server, the set of model data samples is thus has an additional variation on data distribution \cite{2017-DL-TemporalEnsembling}. As a result, we incidentally add noise to the gradient, which raises the variance of the norm of the gradient. This high variance thus increases the gradient stochasticity for the pre-train model \cite{2018-DL-DirectionalSGD}, which improves the model parameters dataset generalization. Therefore, the received dataset is appropriate for the HCFL training phase. Afterward, the model which is obtained at each epoch from the pre-model training phase is utilized for our HCFL model training process. The training process is illustrated in Figures~\ref{fig:HCFL-compressor} and \ref{fig:HCFL-extractor}. In these figures, the HCFL is trained as demonstrated in \eqref{eq:HCFL-update}. As explained in Section~\ref{subsec:problem-formulation}, the trained HCFL model thus has the generalized features of the distributed clients' models in practice.

\textcolor{fix}{
\subsection{Discussion on the compatibility of HCFL with asynchronous FL}
In its simplest version, our proposed HCFL, like the majority of existing FL approaches, suffers from the stragglers in asynchronous FL. Nevertheless, due to the similar structure with the conventional FL process, various resource allocation techniques (e.g., \cite{2021-FL-UserScheduling}, or \cite{2021-FL-EnergyEfficiency}) can be integrated into HCFL to mitigate the straggling phenomenon. we reserve further research on this topic for the future.
}

\section{Performance Analysis on Lossy Compression}
\label{sec:hcfl}
In this section, we theoretically analyze the convergence of HCFL-integrated FL when applied in IoT networks which consist of a large number of devices. We first evaluate the performance of FL by considering its impacts on HCFL. We then prove that in an FL process with a lot of clients in general or in FL-integrated massive IoT network system in particular, HCFL can achieve a performance which is as good as that of the conventional FL process. Furthermore, we analyze the performance of HCFL model and provide more detailed analysis on relationship between the model reconstruction error and three features, including model complexity, compression ratio, and original data entropy. 

\subsection{Convergence Analysis}
\label{sec:Convergence-Analysis}
Using the FedAvg method \cite{mcmahan2017communicationefficient}, we can demonstrate the convergence properties of FL when implementing our proposed HCFL scheme. As the number of clients (i.e., IoT devices) increases, it is guaranteed that aggregated weights are close to the initial desired values.
\begin{figure}[t]
\captionsetup{justification=centering}
\centerline{\includegraphics[width = 0.99\linewidth]{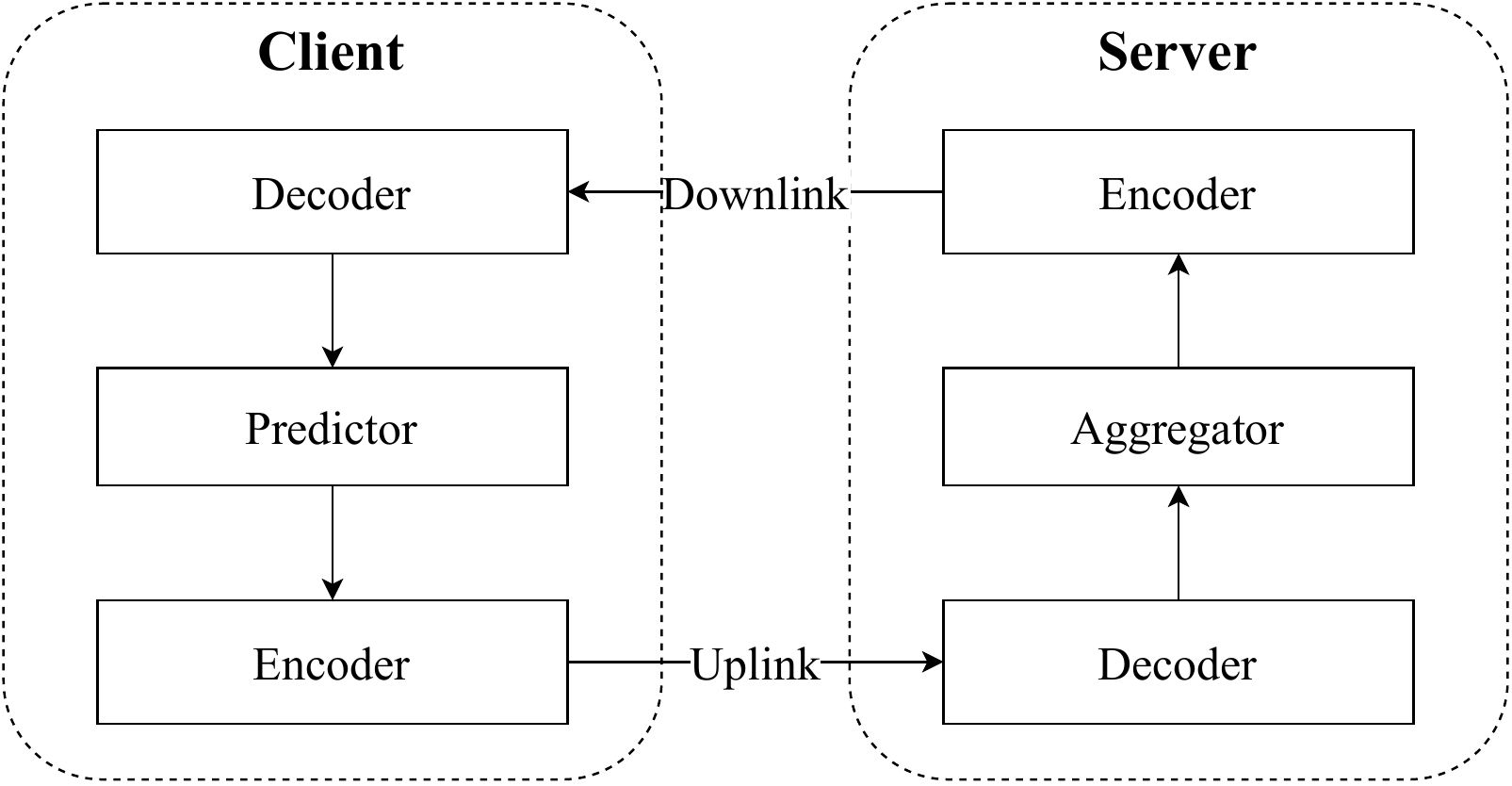}}
\caption{Structure of a general autoencoder with HCFL compression.}
\label{fig2}
\end{figure}
We consider a compression-integrated FL scenario between a single client in a group of selected clients and a server as shown in Fig.~\ref{fig2}. The proposed compression consists of two components: the encoder (which is integrated in the clients) and the decoder which extracts the compressed model sent by the client through the propagation channel. By applying the undercomplete autoencoder \cite{Goodfellow-et-al-2016}, we can compress the original data by reducing its dimensions and important information simultaneously. The ratio of input data to the output data of the encoder, which is equal to the ratio of the output data to the input data of the decoder, represents the compression ratio of the model compressor.
\newline
The autoencoder only extracts useful information on the distribution of the data due to the reduced dimension of the encoded data. The number of faded features increases as we configure the HCFL at a higher compression ratio \cite{Goodfellow-et-al-2016} (the HCFL compression ratio is defined by the proportion between the encoder input size and the encoder output size). Thus, the decrypted data at the output of the decoder in the server always have a reconstruction error rate in comparison with the original data. The problem of autoencoder error and the capability of the compression in FL is simplified by converting original form of the autoencoder's representation into another close-form.
\textcolor{fix}{\begin{theorem} Given the model compression distortion rate $\mathcal{L}(w)$, the uncertainty that the geometrical distance between aggregated model from lossy compression $w_t$ and ideal aggregated model $\widetilde{w}_t$ is inversely proportional to number of user $K$.
\begin{equation}\label{CERTAINTY2}
\begin{gathered}
P\Big(\abs{w_t - \widetilde{w}_t} \geq \alpha\Big) \leq \frac{2}{(K\alpha)^{2}}\mathcal{L}(w).
\end{gathered}
\end{equation}
\label{theorem:FL-distortion-rate}
\end{theorem}
\textit{Proof.} The proof is demonstrated in Appendix~\ref{appendix:proof-on-theorem-FL-distortion-rate}.}

\textcolor{fix}{The theorem~\ref{theorem:FL-distortion-rate} is the proof of convergence of implementation of the aggregations algorithm at the server when the number of clients in the federated network is large enough. Despite the high loss value from the autoencoder, we can achieve the aggregated model, which is close to the ideal value of the model weights without the HCFL compression.} For instance, the uncertainty of an event when $\mathcal{L}(w) = 2.5$, an expected loss of the aggregated weights compared to the client weights is $\alpha = 0.01$, and $K = 10000$ FL devices, is: \(P\Big(\abs{w_t - \widetilde{w}_t} \geq \alpha\Big) \leq \frac{2}{(10000*0.01)^{2}}2.5 = 0.0005\). Therefore, the certainty is equal to \(99.95\%\). 
According to \cite{de2021survey}, the number of IoT and mobile devices with MEC capabilities will be extremely large in beyond 5G and future 6G systems, i.e., $24$ billion devices by 2030 \cite{de2021survey}. As a result, the certainty induced by the proposed HCFL system can be further reduced in emerging scenarios with massive IoT connectivities, showing the efficiency and great potential of the HCFL system.
\newline
On the one hand, we prove that the distortion rate induced by the HCFL's lossy compression process is decreased significantly by the FL's number of participated clients. On the other hand, the distortion rate from the lossy compression also encourages FL models in distributed devices to execute the exploration steps toward gradient descent. This noise acts similarly with the gradient noise, which is proposed in \cite{2015-DL-AddingGradientNoise}.
\section{Impact of model complexity to HCFL settings}
\label{sec:HCFL-settings}
\begin{figure}[t]
\centerline{\includegraphics[width=0.95\linewidth]{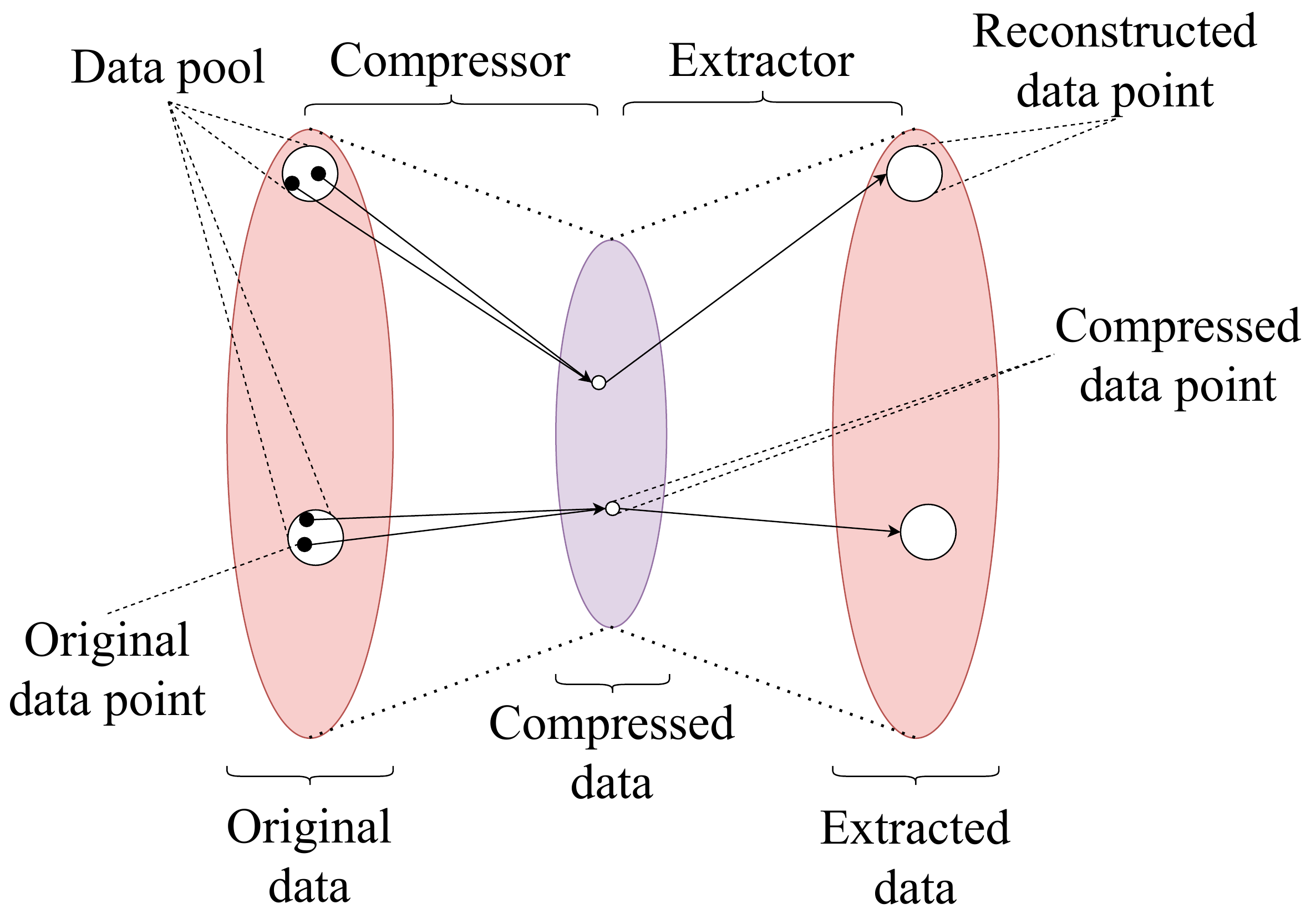}}
\caption{Visualization of mutual information between original data and extracted data in HCFL.}
\label{hcfl-mutual-information}
\end{figure}

In the HCFL system, the encoder aims to extract the representative features of the original data. There is a direct mapping through the neural network from each individual particle from the primordial model to the compressed output at the encoder. The compressed data is then mapped straightforwardly to the extracted output at the decoder. The procedure can be described in Fig.~\ref{hcfl-mutual-information}. 
As we can see from the figure, there is a representative vector for each pool of original data points that have the same characteristics. The reconstructed vector is sampled from the previously mentioned pool and can represent a group of data points in that pool. Hence, the compressed data is the mutual information between the original and the reconstructed data. The lower the compression ratio is, the more information that the compressed data transfers from the original data to the extractor. Therefore, the fewer the number of original data features that one compressed data component needs to represent, the better the HCFL can perform.
\textcolor{fix}{\begin{theorem} The reconstruction loss $\mathcal{L}(w)$ of HCFL is proportional to 
\begin{align}\label{eq:model-complexity-analysis}
    \mathcal{L}(w) &\approx \frac{H(W)-H(C)}{N\log{(2\pi{e})}} \notag\\
    &= \frac{\sum_{N}P(w)\log{P(w)}-\sum_M{P(c)\log{P(c)}}}{N\log{(2\pi{e})}},
\end{align}
where the distribution of $C$ follows the joint distribution $P(X,g^1,\dots,g^\mathcal{V})$ with $\mathcal{V}$ being the depth of the compressor's hidden layer. Thus, we have
\begin{equation}\label{AppendixA-5}
\begin{split}
P(c) &= \prod^{\mathcal{V}+1}_{i=1}{P(g^i|g^{i+1})} \\
&= \prod^{\mathcal{V}}_{i=1}{\prod^{n_i}_{j=1}{P(g^i_j|g^{i+1})}}{\prod^{M}_{k=1}{P(c_j|g^{\mathcal{V}})}} \\
&= P(g^1,\dots,g^{\mathcal{V}}){\prod^{M}_{k=1}{P(c_j|g^{\mathcal{V}})}}, 
\end{split}
\end{equation}
where $P(g^1,\dots,g^{\mathcal{V}})$ is the joint probability of the $V$ hidden layers neural network of the compressor and the factorized conditional distribution of the compressed output data. 
\label{theorem:FL-model-complexity}
\end{theorem}
\textit{Proof.} The proof is demonstrated in Appendix~\ref{appendix:proof-on-theorem-model-complexity}.}

The two equations (\ref{eq:model-complexity-analysis}) and (\ref{AppendixA-5}) give us a prerequisite for the convergence of the HCFL compression. To achieve the HCFL compression system with low restoration error, the HCFL network's complexity and compression ratio need to be taken into account. When the HCFL with higher compression ratio and more complicated input data is applied, the neural network model is needed to be the deeper. Moreover, the data segmentation technique is considered as a robust method when dealing with models with a huge size and tremendously complex structure.

\section{Performance Evaluation} 
\label{sec:simulation}
In this part, we evaluate the performance of the proposed HCFL compression. Depending on different testing scenarios, we establish multiple settings to examine the HCFL under distinct conditions in Section~\ref{V-1}. We then evaluate the HCFL compression efficiency with various compression ratios in Section~\ref{V-2}. Our experiments show that HCFL can achieve up to 32 times of reduction in data size. Then, we evaluate the impact of clients number on FL convergence rate in Section~\ref{V-3}. Section~\ref{V-4} shows the influence of the client's predictor hyper-parameters on the HCFL-integrated FL to analyze the suitable setting when applying HCFL on FL.
\subsection{Settings}
\label{V-1}
The reconstruction error of the HCFL-compression framework is considered along with the losses of the predictors at the clients' ends with the updated model parameters under different FL settings to simplify the evaluation. For the simulation, we use a set of $100$ clients in synchronous FL system for assessments. A comprehensive description of the setting is presented below. 

\textbf{Evaluation on communication efficiency:} As mentioned in Section~\ref{section:2}, our proposed network model is in compliance with the 3GPP standard. To be more specific, our proposed HCFL is located as the effectiveness encoder in \cite[Figure~3a]{2021-Comm-WhatisSemCom}. Therefore, any package error is pre-processed and corrected via HARQ protocol \cite{3GPP_38300}. Therefore, the encoded data from HCFL is guaranteed to be flawless and independent of the channel model. As a result, we only consider the data compression ratio as a communication efficiency evaluation metric in this research.

\textbf{Dataset:} We choose two popular benchmark datasets that are widely used for classification. We focus on the performance of FL which is not affected by the non-IID data, where each client holds a separated dataset from the other.
\begin{itemize}
    \item \textit{MNIST} \cite{lecun-mnisthandwrittendigit-2010}: consists of $60,000$ training and $10,000$ testing gray-scale images of ten classes of handwritten digits. Each image has the dimension of $28 \times 28$ pixels. For more details, there are $100$ clients in the system where each client has $600$ samples that are independent and identical with each other. Because of the simplicity of MNIST, this dataset is mostly used to train small networks. 
    \item \textit{EMNIST} \cite{DBLP:journals/corr/CohenATS17}: includes $131,600$ images of $47$ balanced classes with a size of $28 \times 28$ pixels. The dataset is partitioned into $100$ independent and identical clusters of $11,280$ images, each is represented for the dataset of each different client.
\end{itemize}

\begin{table}[t]
\centering 
{\color{firstreview}
\renewcommand{\arraystretch}{1.25}
\caption{Performance of \textcolor{firstreview}{HCFL and other compression methods} on LeNet-5 model training with MNIST dataset with different compressing ratio (Communication Cost in $100$ rounds on IID Data). Each round, ten out of $100$ clients are participated in the training ($C = 0.1$).} 

\begin{tabular}{| c | c | c | c |} 
\hline
Compress  & Reconstruction  & Encoded Size & True \\ [0.5ex] 
Method & error & Up/Download & Compress\\
& & (MB) & Ratio \\
\hline \hline  
FedAvg & $0.0$ & $20500$/$20500$ & $1.000$\\ 
T-FedAvg & N/A & $1281$/$1281$ & $15.999$ \\ 
HCFL 1:4 & $0.0016$ & $5170$/$5170$ & $3.965$ \\ 
HCFL 1:8 & $0.0037$ & $2620$/$2620$ & $7.824$ \\  
HCFL 1:16 & $0.0037$ & $1370$/$1370$ & $14.963$ \\
\textbf{HCFL 1:32} & {$\boldsymbol{0.0040}$} & $\boldsymbol{757}$/$\boldsymbol{757}$ & $\boldsymbol{27.080}$ \\ 
\hline 
\end{tabular}
\label{tab:compression} 
}
\end{table}

\textbf{Models:} In order to assess the performance of HCFL, we use two deep learning models: LeNet-5 and 5-CNN, which represent two popular convolutional deep network architectures. The configuration is described in detail as follows:
    \begin{itemize}
        \item LeNet-5 \cite{726791} is a simple convolutional neural network. It contains the basic units of a convolutional neural network. In LeNet-5, in which the data is first feed-forward through two sets of feature mapping. Each set includes one convolutional layer, with the max-pooling layer is followed subsequently. The two proposed sets are connected sequentially, followed by a fully connected layer that uses ReLU as their main activation function. A ten-node group of softmax is applied at the output layer in order to execute the probability classification for the model.
        \item 5-CNN is an advanced deep neural network. \textcolor{duongfix}{It comprises two main components: the first component includes five convolutional layers, and the second consists of two fully connected layers.} A max-pooling layer is used after each convolutional layer, and a ReLU activation function is applied at the end of each max-pooling layer. Then, additional fully connected layer is applied to process the classification works. Dropout layers are added, followed by each fully connected layer to reduce the impact of overfitting.
    \end{itemize}

\textbf{Dataset segmentation:} The model parameter datasets of the FL model in our simulations are processed as follows:
\begin{itemize}
    \item \textit{MNIST}: The convolutional layers and dense layers are trained in different HCFL compressors and extractors. Each of the HCFL learns the different distribution of each group of convolution kernel parameters or fully connected weights. Hence, we can achieve the high compression efficiency for the HCFL. 
    \item \textit{EMNIST}: Due to the complexity of the 5-CNN model parameters, we fractionate the dense layers' parameters into 8 balanced parts in order to reduce the entropy of each part. \textcolor{duongfix}{The HCFL compressor needs to execute eight different trainings} and the HCFL compressor setting is stored in the HCFL memory. Whenever the FL model applies the HCFL, the particular HCFL setfting is loaded from the memory to process the corresponding segmented dataset.  
    \end{itemize}

\textbf{Initial implementation details:} Each round, the server selects $10$ customers from a pool of $100$ clients at random. We will use a fixed learning rate of $0.01$ as an example. The number of epochs at each client is set to five, and the mini-batch size is set to $64$. This implementation is applied to the experiment in Section~\ref{V-2}. In Sections~\ref{V-3} and \ref{V-4}, the numbers of participating clients $K$, local epoch $E$ and local batch-size $B$ are applied with different values to evaluate the HCFL under different settings. The detailed settings of those experiments are demonstrated in sections~\ref{V-3} and \ref{V-4}.

\begin{table}[t]
\renewcommand{\arraystretch}{1.25}
\centering
{\color{firstreview}
\caption{Performance of HCFL compression on 5-CNN model training with EMNIST dataset with different compressing ratio (Communication Cost in 100 rounds on IID Data). Each round, $100$ out of $100$ clients are participated in the training ($C = 0.1$).} 
\centering 
\begin{tabular}{| c | c | c | c |} 
\hline
Compress  & Reconstruction  & Encoded Size & True \\ [0.5ex] 
Method & error & Up/Download & Compress\\
& & (MB) & Ratio \\
\hline \hline  
FedAvg & 0.0 & 27200/27200 & 1.000\\ 
T-FedAvg & N/A & 1,650/1,650 & 16.485\\ 
HCFL 1:4 & 0.0486  & 5630/5630 & 4.831 \\ 
HCFL 1:8 & 0.0495 & 2930/2930 & 9.283 \\  
\textbf{HCFL 1:16} & \textbf{0.0501} & \textbf{1570/1570} & \textbf{17.324} \\
\textbf{HCFL 1:32} & \textbf{0.0693} & \textbf{910.55/910.55} & \textbf{29.872} \\ 
\hline 
\end{tabular}
\label{tab:compression-emnist} 
}
\end{table}

\subsection{\textcolor{firstreview}{Robustness to Compression Proficiency}}\label{V-2}

We first perform some experiments with the HCFL compression only to validate the affection of the compression ratio on the reconstruction error of the extracted data. We evaluate the algorithm after $100$ training rounds with an equal number of training samples for each client, and every client is computed in $10$ independent runs each round for a fair comparison. In this subsection, we consider three main compression efficiency features, including compression efficiency, computational delay, and HCFL-assisted FL accuracy.

From the perspective of compression efficiency, we evaluate the actual compression performance when applying different compression ratio settings in HCFL on FL. Tables~I and II show the reconstruction error \textcolor{firstreview}{and the compression efficiency of HCFL and the benchmarks, including FedAvg \cite{mcmahan2017communicationefficient} and T-FedAvg \cite{xu2020ternary}, respectively, under different compression ratio settings. We apply the benchmarks on two dataset: MNIST and EMNIST.} Assume that there are $10$ users participating in each round over $100$ rounds, the total capacity which is necessary for the transmission between clients and the centralized server is $20.5$~GB for the LeNet-5 model and $27.2$~GB for the 5-CNN model, respectively. This large amount of data is an enormous burden on large-scale IoT networks. The communication cost is expected to be more significant in high complexity deep networks (ie., AlexNet \cite{10.5555/2999134.2999257} or ResNet \cite{DBLP:journals/corr/HeZRS15}). The relationship between the model parameter size and the communication time is given as 
\begin{equation}\label{COMMUNICATION_TIME}
\begin{split} 
T^{comm}_{k} = {s_k}/{{R_{k}}},
\end{split}
\end{equation}
where $s_k$ is the model parameter size and $R_{k}$ is the transmission rate. From~\eqref{COMMUNICATION_TIME}, by reducing transmitting model parameter size, we can reduce the communication time with the same ratio. It is worth noticing that when we increase the compression ratio, the communication efficiency is improved. In case HCFL-assisted FL is applied, we can reduce a huge amount of data on the transmission link and the required communication time as well. In particular, there are four proposed compression ratios in our work: $1$:$4$, $1$:$8$, $1$:$16$, and $1$:$32$. The data from the two tables reveals that applying the setting of $1$:$4$ helps to reduce the communication cost to $25.22\%$ and $20.7\%$ when working on LeNet-5 model and 5-CNN. Especially, by applying the $1$:$32$ setting, the communication cost can be reduced to $3.7\%$ (from $20.5$~GB in the conventional FL to $757$~MB in HCFL-assisted FL) in LeNet-5-integrated clients and $3.35\%$ (from $27.2$~GB in the conventional FL to $910.55$~MB in HCFL-assisted FL) in 5-CNN-integrated clients. \textcolor{firstreview}{Furthermore, 
it can be observed from Table~\ref{tab:compression} that HCFL at compression ratio of $1$:$32$ outperforms the T-FedAvg \cite{xu2020ternary} in communication efficiency as the T-FedAvg's maximum compression ratio is capped at $16$ times.}

\begin{figure}[t]
\centerline{\includegraphics[width=0.85\linewidth]{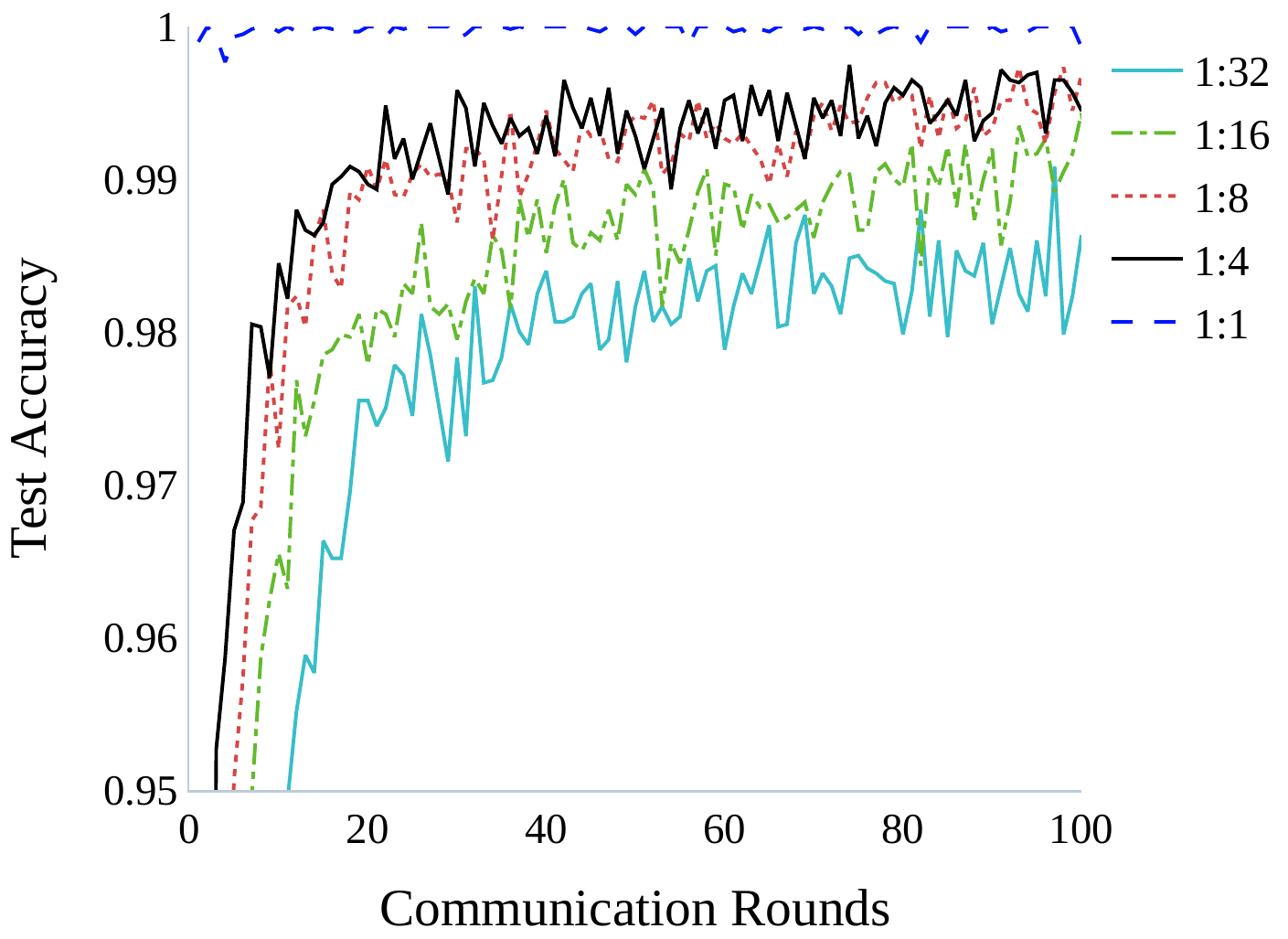}}
\caption{Aggregation accuracy of HCFL on MNIST dataset at different compression ratio settings.}
\label{compression_ratio_affection}
\end{figure}

\begin{table}[t]
\renewcommand{\arraystretch}{1.25}
\caption{Computational delay of HCFL-assisted FL on LeNet-5 model training with MNIST and on 5-CNN model training with EMNIST (Evaluation is averaged in 100 rounds of FL process). Each round, ten out of $100$ clients are participated in the training ($C = 0.1$).} 
\centering 
\begin{tabular}{|c|c|c|c|c|} 
\hline
\multirow{3}{*}{\begin{tabular}[c]{@{}c@{}}Compression \\ Ratio\end{tabular}} & \multicolumn{4}{c|}{Computational time (second)}                                                       \\ \cline{2-5} 
                                                                              & \multicolumn{2}{c|}{LeNet-5-integrated client} & \multicolumn{2}{c|}{5-CNN-integrated client} \\ \cline{2-5} 
                                                                              & client                 & server                & client                & server               \\ \hline
Baseline                                                                      & 2.133                  &0.0142                 & 2.171                 &0.0228                \\ \hline
1:4                                                                           & 2.146                  &0.1095                 & 2.190                 &0.2353                \\ \hline
1:8                                                                           & 2.178                  &0.1136                 & 2.192                 &0.2793                \\ \hline
1:16                                                                          & 2.183                  &0.1143                 & 2.195                 &0.3130                \\ \hline
1:32                                                                          & 2.283                  &0.1231                 & 2.209                 &0.3435                \\ \hline
                                                                              
\end{tabular}
\label{tab:computational_delay} 
\end{table}

From the perspective of computational delay, we assess the average computational time for both clients and server in a long-term FL process. Because the HCFL components are deployed in both server and clients, we conduct the delay calculations on both centralized server and distributed clients. We calculate the average delay value over $100$ communication rounds in order to get the fair assessment. The computation delay is demonstrated in Table~\ref{tab:computational_delay}. As observed in the table, the assessment is implemented on both LeNet-5-integrated client (MNIST dataset) and 5-CNN-integrated client (EMNIST dataset). The HCFL process time can be calculated as follows:
\begin{equation}\label{COMPUTATION_TIME}
\begin{split} 
T^{HCFL}_{r} = T^{comp}_{r} - T^{comp}_{1},\end{split}
\end{equation}
where $r$ is the compression ratio of the HCFL system, $T^{HCFL}_r$ is the HCFL process computation delay and $T^{comp}_r$ is the client's total computation delay with the compression ratio $r$. From the table, we can see that the HCFL process on both clients and server is low (less than $40$ milliseconds on client and $350$ milliseconds on server). This time amount is much lower than the predicting process delay on the client (around $2.1$ to $2.2$ seconds).

In the following evaluation of HCFL compression efficiency, we consider the HCFL-assisted FL accuracy. The assessment uses the same setting with ten clients randomly selected from 100 clients in total, the selected client ratio is set at $0.1$, and the batch size is set to the maximum possible value (i.e., equal to the data size at each client) with the number of training epoch set at five. To compare the performance between different compression ratios of HCFL, we train different neural network models for the encoder and decoder of each particular HCFL with the determined compression ratio and then embed them into the recommend FL model. Fig.~\ref{compression_ratio_affection} shows the predicting accuracy at clients with different compression ratio settings at the HCFL compressor. The test accuracy is calculated by the percentage of data in the test set with the predicting label matching their original label. As we can see from the figure, the accuracy of HCFL at the beginning of the distributed training session is relatively low due to the internal error of the data generated by the HCFL compressor when working on MNIST dataset with LeNet-5 as the predicting model. However, the performance of the HCFL-integrated FL can be converged after only six to seven communication rounds. Moreover, although the global accuracy is decreased at the compression rate of $16$ and $32$, the test accuracy at $98\%$ for $1$:$32$ compression rate is at the acceptable threshold. Alternatively, this decline may be attributed to the fact that the more representation of the data vanishes from the original information, the more error-prone that the HCFL-assisted IoT system is exposed to. 

\begin{figure}[t]
\centerline{\includegraphics[width=0.85\linewidth]{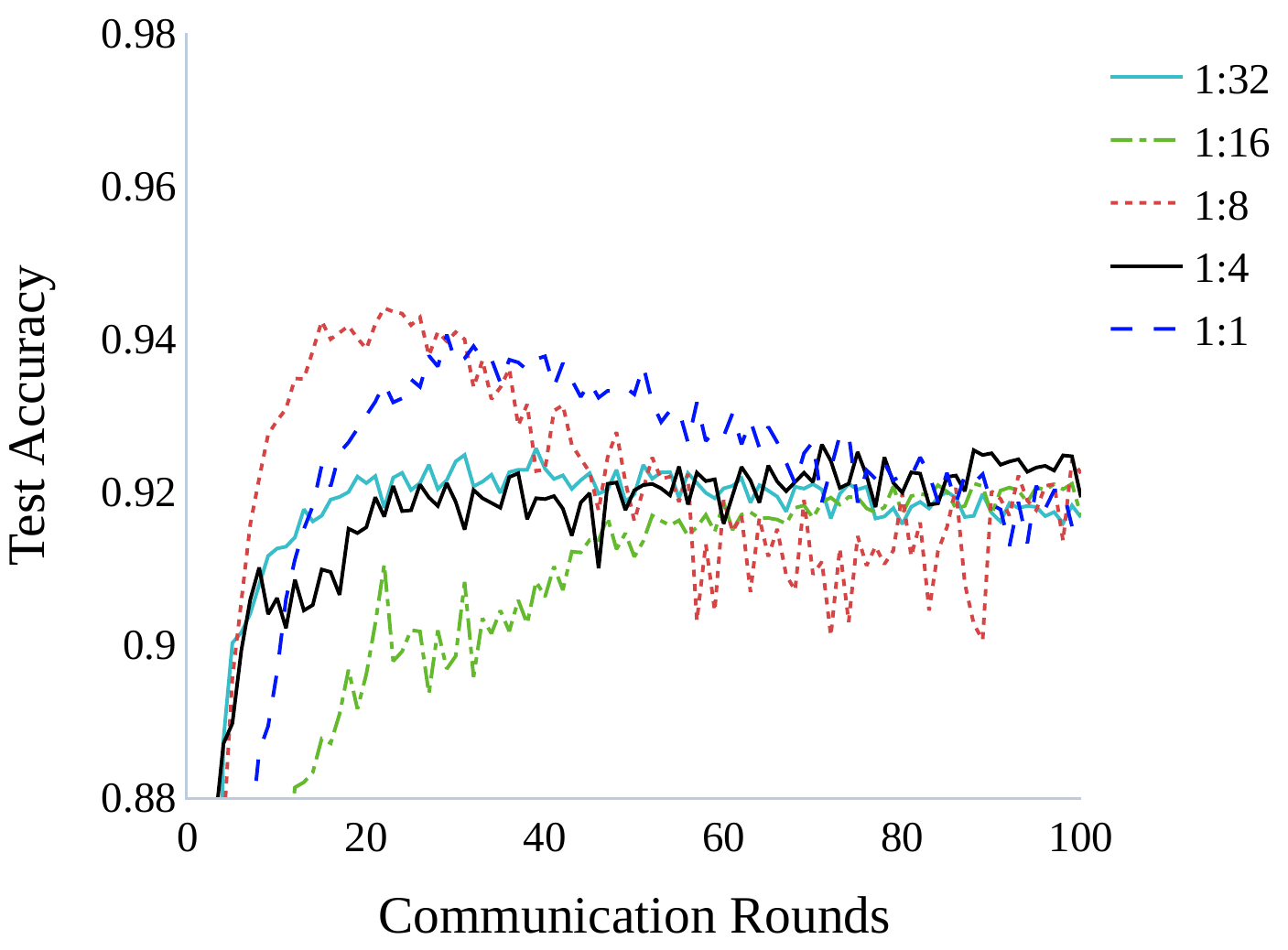}}
\caption{Aggregation accuracy of HCFL on EMNIST dataset at different compression ratio settings.}
\label{emnist_compression_ratio_affection}
\end{figure}

When dealing with such model with higher complexity as 5-CNN for the EMNIST, we need to apply the pre-mentioned dataset segmentation technique for the model parameters. The performance of HCFL on the proposed model is demonstrated in Fig.~\ref{emnist_compression_ratio_affection}. As observed from the figure, due to the random shuffle and division of the full EMNIST dataset for the clients and the stochastic initiation of the client predicting model, the test accuracy of all five cases shows the high fluctuation and different convergence speeds at the primitive stage of the HCFL training. However, the HCFL system can converge within after less than $100$ communication rounds, regardless the proposed compression ratio. The high performance of different compression schemes in the prolonged term is believed to attain on account of the reduction in entropy of the model parameters which is proposed in Section~\ref{sec:HCFL-settings}.

\begin{figure*}[htb]
	\centering
	{\color{firstreview}
	\subfloat[\label{fig:client_number_affection}]{\includegraphics[width=0.35\linewidth]{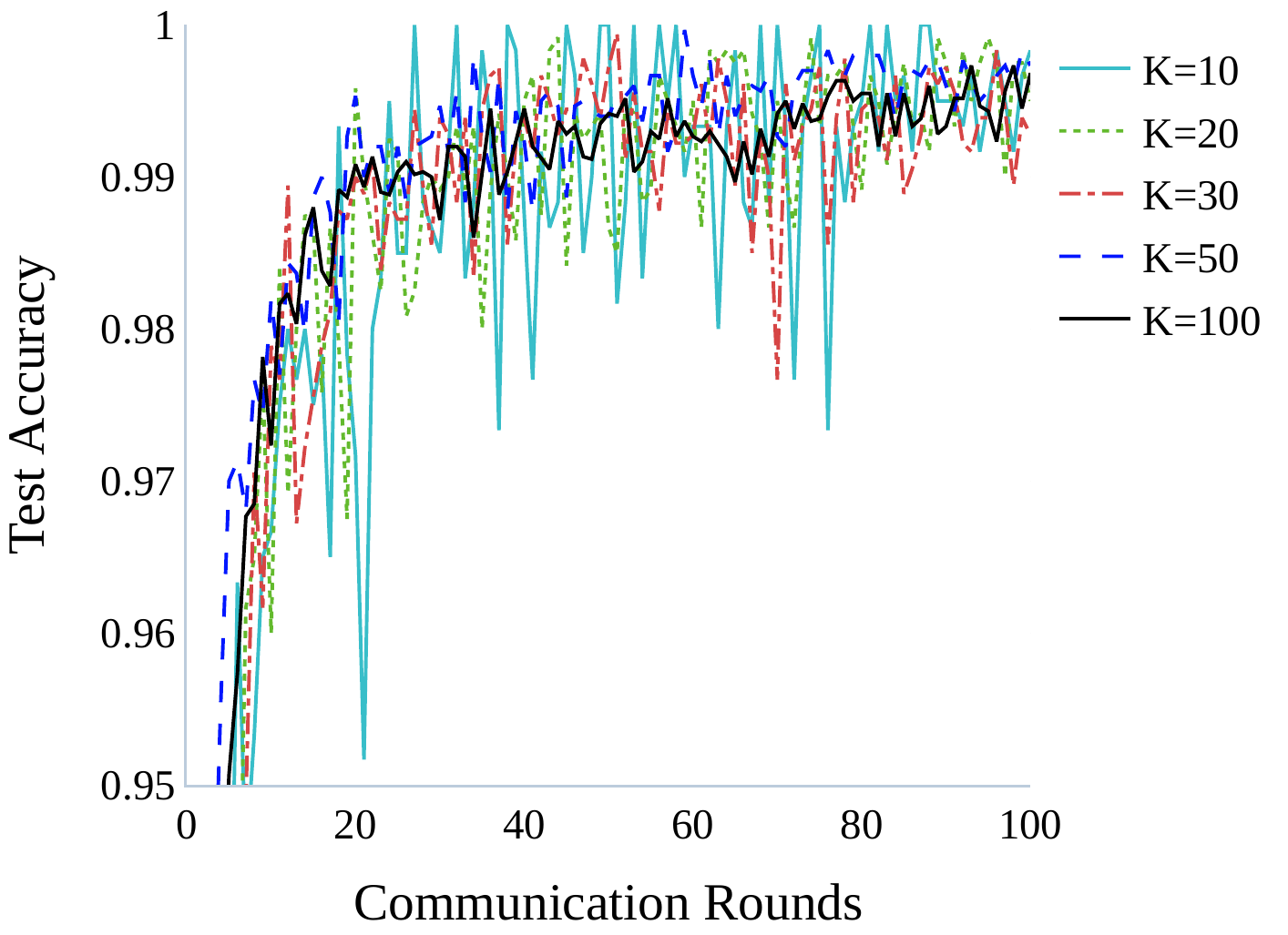}} 
	\quad
	\subfloat[\label{fig:client_number_affection_emnist}]{\includegraphics[width=0.35\linewidth]{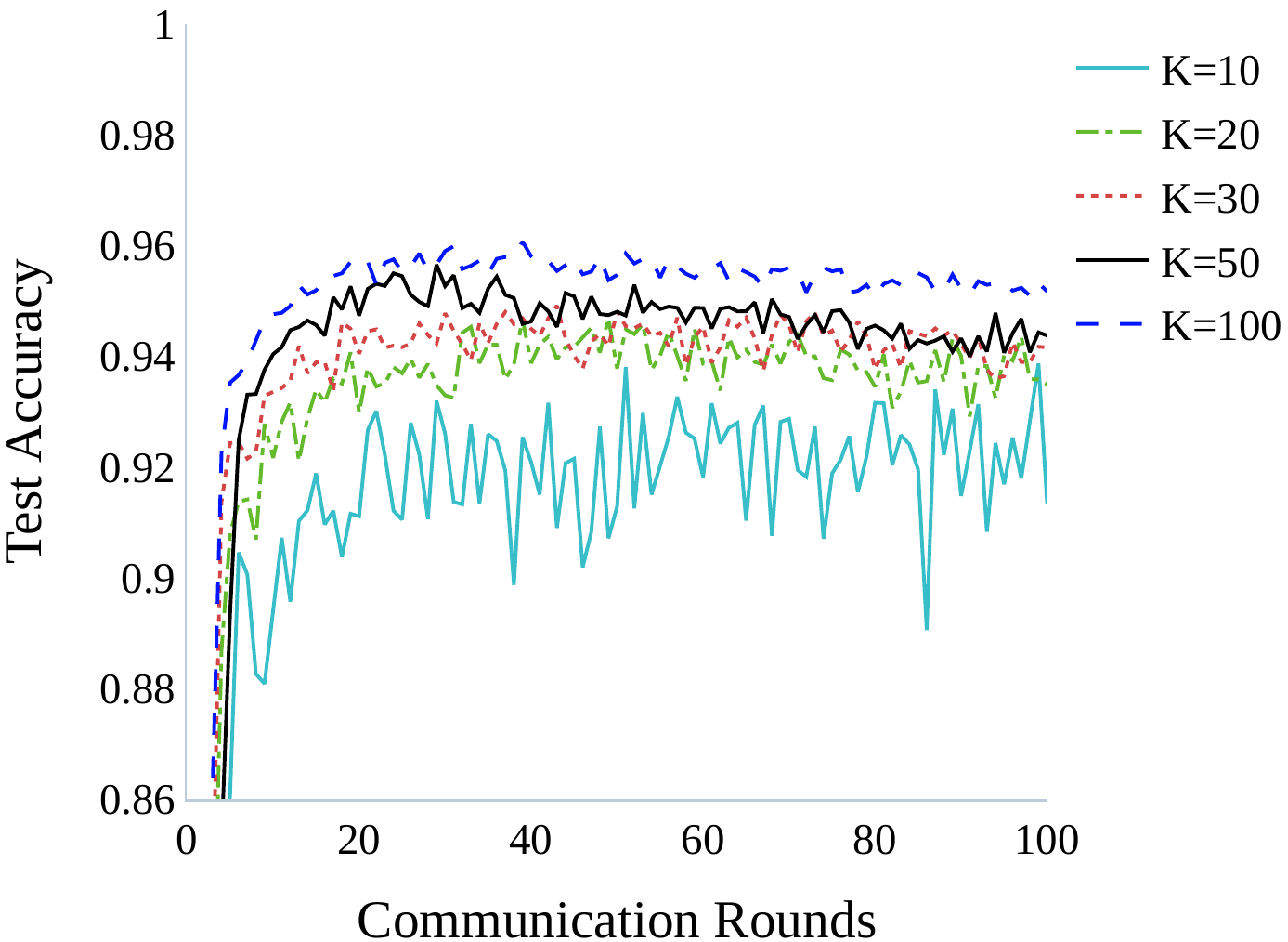}}
    \caption{Implementations of the FL process applying HCFL for MNIST and EMNIST handwritten digit dataset. In these simulations, we set up the FL with different numbers of clients. (a) Number of clients affects the aggregation accuracy of the FL process on MNIST dataset. (b) Number of clients affects the aggregation accuracy of the FL process on EMNIST dataset.}}
\end{figure*}


\begin{figure*}[htb]
	\centering
	\subfloat[\label{fig:HCFL-epoch-accuracy}]{\includegraphics[width=0.35\linewidth]{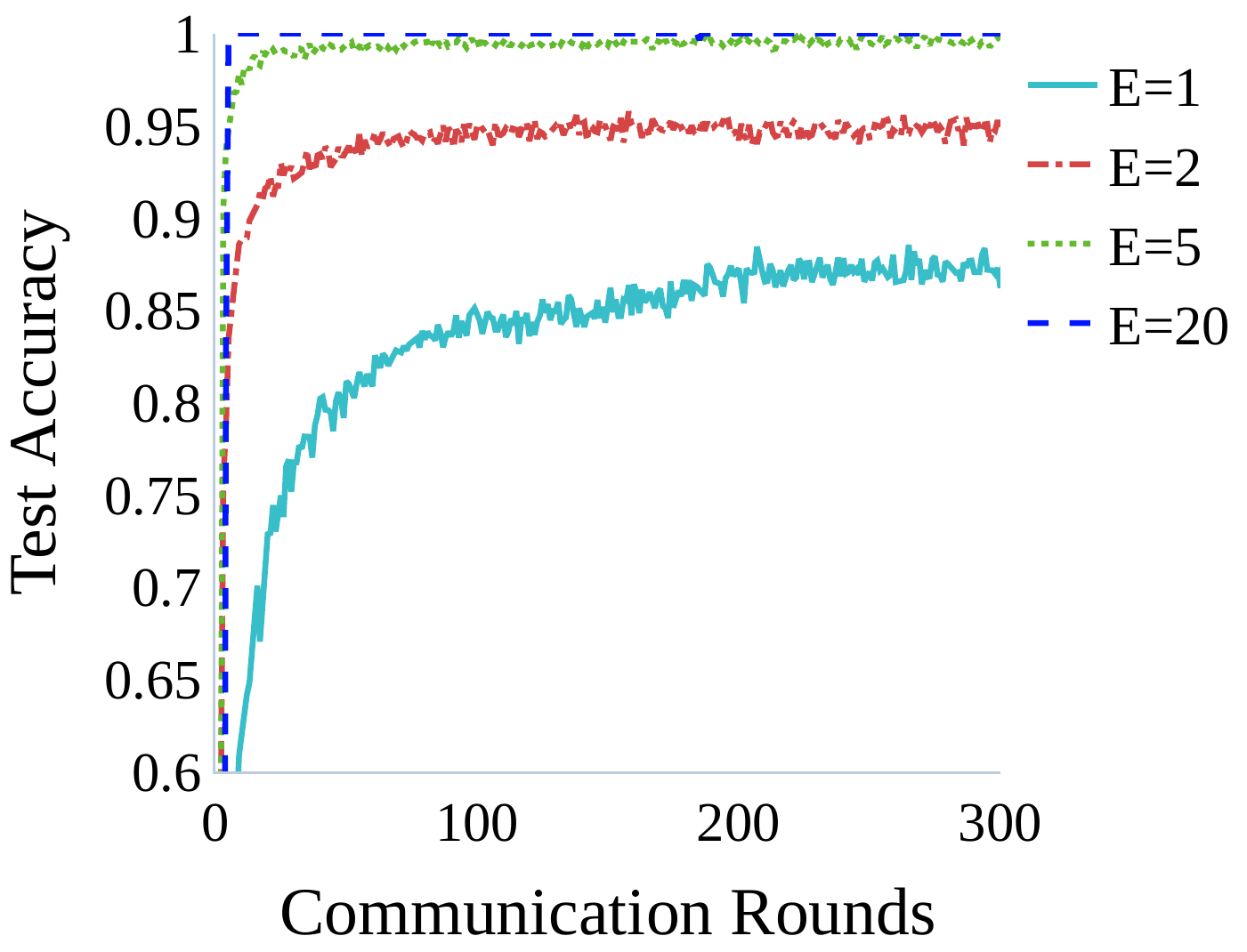}}\qquad
	\subfloat[\label{fig:HCFL-epoch-loss}]{\includegraphics[width=0.35\linewidth]{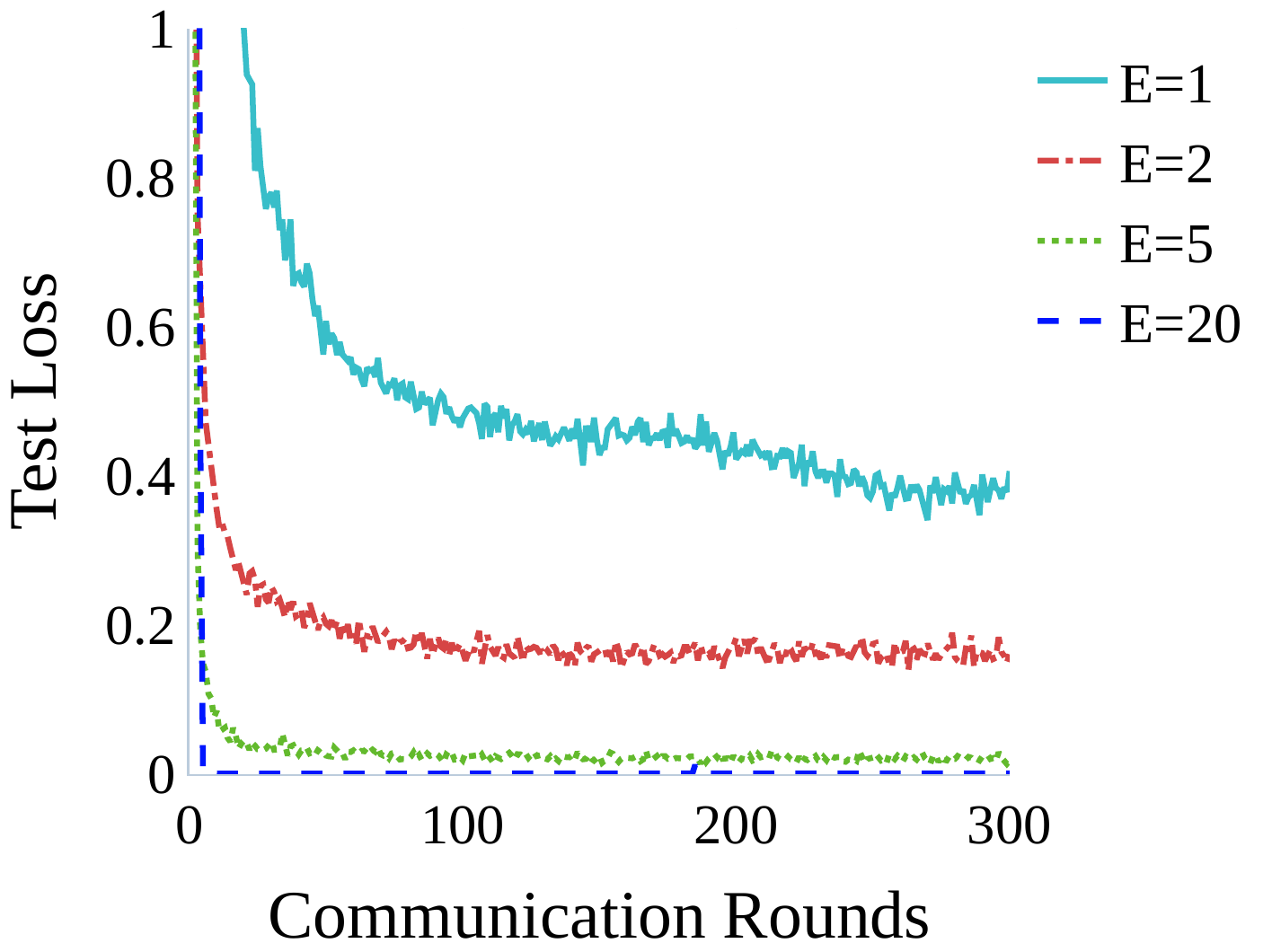}} 
	\caption{Aggregation accuracy on LeNet-5 model with MNIST dataset for different local epochs ($E$). Number of total clients and participated ratio are fixed ($K=100, C=0.1$).}
\end{figure*}

\begin{figure*}[htb]
	\centering
	\subfloat[\label{fig:HCFL-batch-accuracy}]{\includegraphics[width=0.35\linewidth]{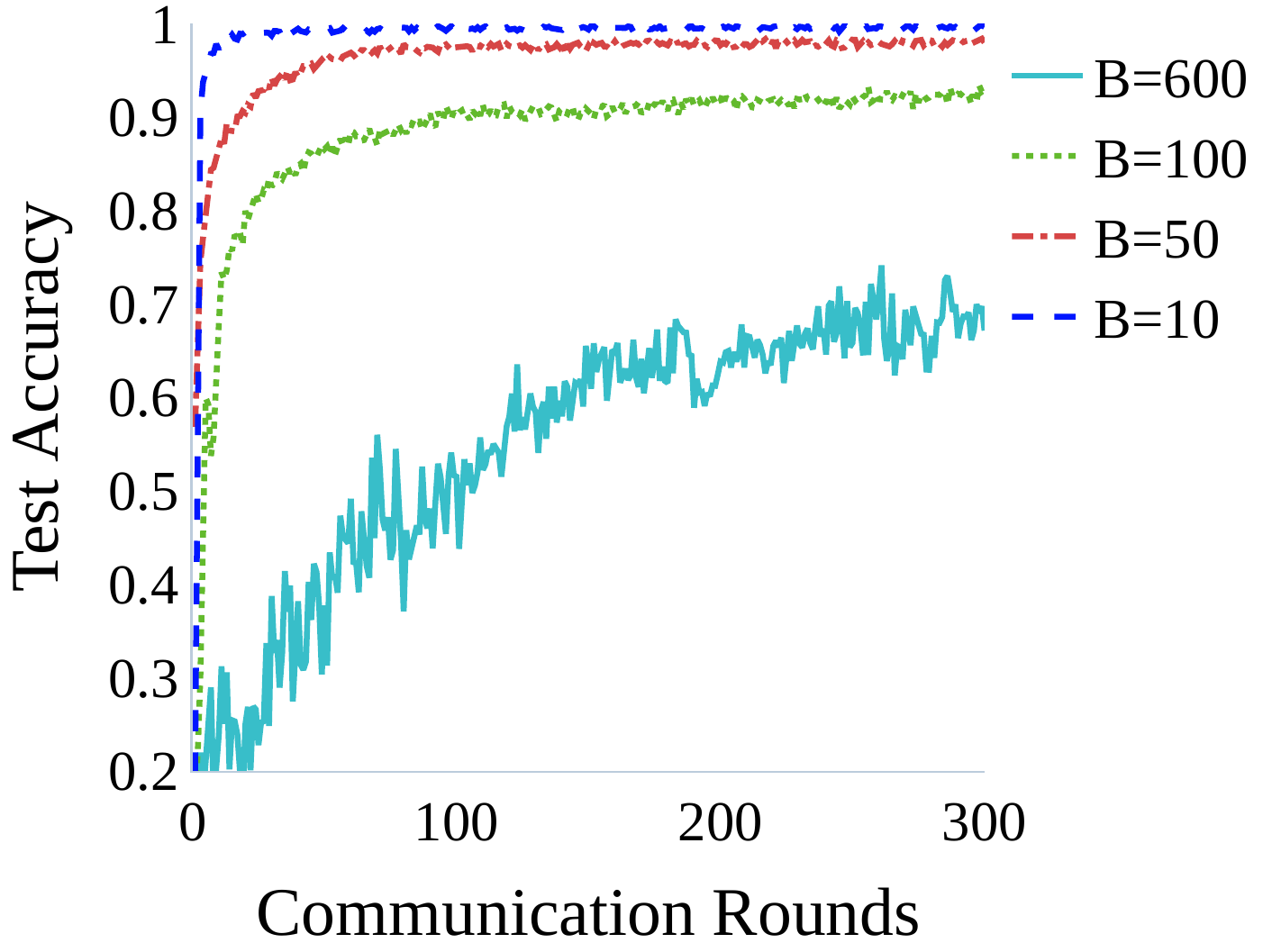}}\qquad
	\subfloat[\label{fig:HCFL-batch-loss}]{\includegraphics[width=0.35\linewidth]{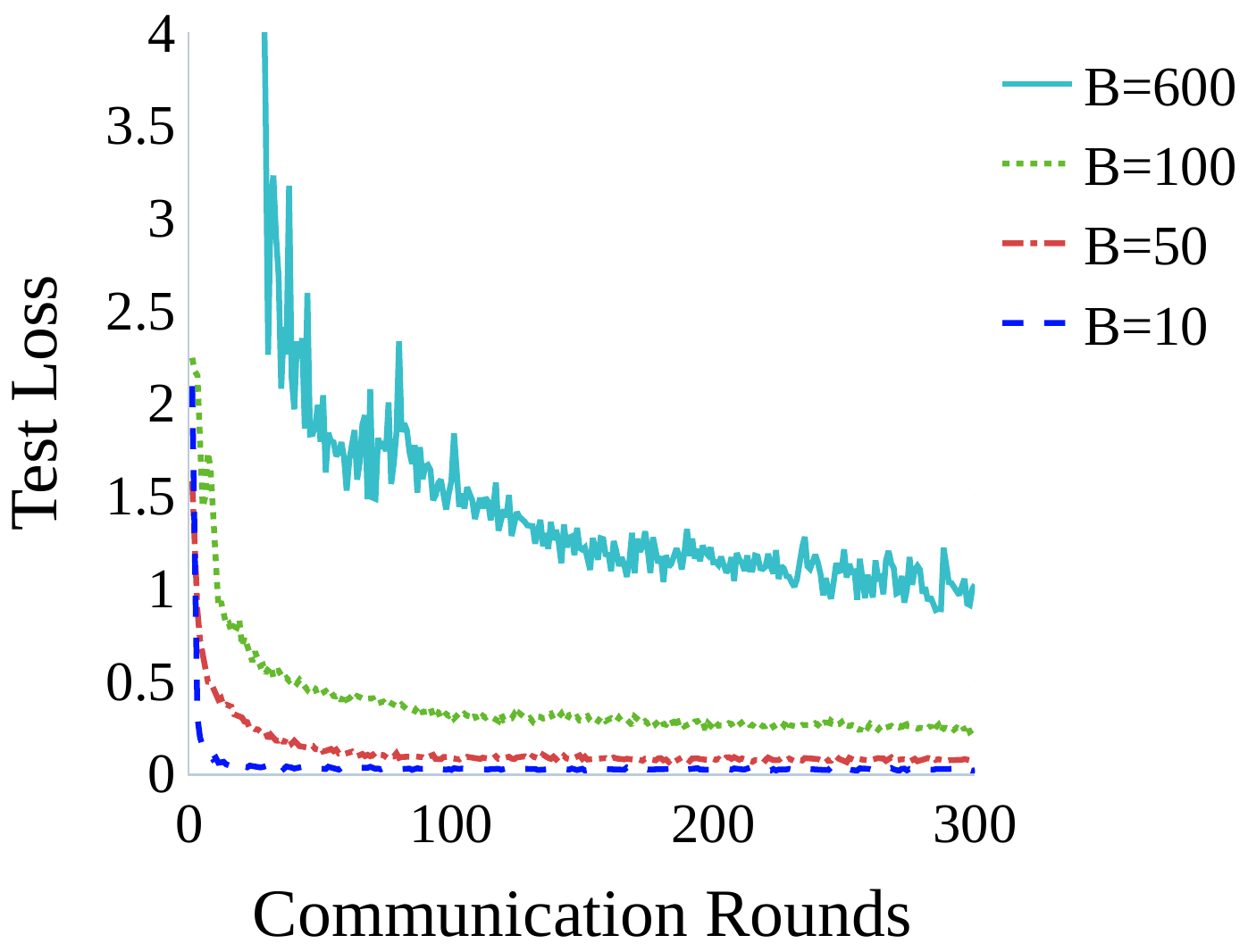}} 
	\caption{Aggregation accuracy on LeNet-5 model with MNIST dataset for different batch size ($B$). Number of total clients and participated ratio are fixed ($K=100, C=0.1$), number of epoch ($E=5$).}
\end{figure*}

\subsection{\textcolor{firstreview}{Contribution of Participating Client Quantity to Global Convergence}} \label{V-3}
In this assessment, we evaluate the impact of the number of participating clients on the convergence of the error-prone compressed data. The HCFL compression is applied in FL with a dissimilar quantity of clients. We evaluate them on two datasets, i.e., MNIST and EMNIST. The results obtained by the HCFL-assisted FL on MNIST and EMNIST are shown in Fig.~\ref{fig:client_number_affection} and Fig.~\ref{fig:client_number_affection_emnist}, respectively. In general, it is shown that the performance of the HCFL compression function in the FL model can attain expected efficiency with a various number of clients participating in the FL model. However, the larger the number of clients is, the sooner the system will converge to the global extrema. Therefore, with a large number of clients using HCFL, predictors accuracy can quickly achieve a high performance, and thus the accuracy will be more stable through communication rounds. For example, for the setting $K = 100$, the test accuracy of the LeNet-5 model on MNIST dataset reaches 99\% after less than $20$ communication rounds with a low standard deviation (less than $1\%$). In contrast, the system with a low number of assigned clients ($K = 10$) has a relatively high standard deviation of accuracy (more than $3\%$) after $80$ communication rounds.

In particular, different deep network models being used on the clients make distinct impacts on the whole HCFL-assisted FL. For instance, the complex model 5-CNN with the EMNIST dataset shows a more significant outcome than the mentioned evaluation on the MNIST dataset. For the system with $100$ clients in comparison with the $10$-client system, the test accuracy is higher with a notably lower variance. \textcolor{duongfix}{Generally, in a predetermined number of training rounds,  the number of clients affects the FL process accuracy and convergence speed. As more clients participating in the FL process each round, the model's absolute precision and training speed suffer from fewer adverse effects.} Nevertheless, once $K$ is improved to a particular level, the advancement of the system performance will be less noteworthy, and sometimes it even sets about degrading.
When bringing the FL process into practice, we can face difficulties that as $K$ expands, more and more clients update their local parameters to the server. As a consequence, the communication and computation cost of the FL model amplify significantly. Fortunately, as we can see in the simulation results in Fig.~\ref{fig:client_number_affection} and Fig.~\ref{fig:client_number_affection_emnist}, there always exists an upper bound of performance saturation when $K$ increases. \textcolor{duongfix}{This is encouraging because in real-world applications with large scale of clients, we only need to select a set of clients from the network to execute the FL process in each communication round. This procedure     save a considerable amount of communication cost for the FL process.}


\subsection{Contribution of Distributed Model Hyper-parameters to Global Convergence}\label{V-4}
In this part, we investigate the impact of compression reconstruction error on different FL process settings. Two client model's hyper-parameters that we consider are: Epoch and batch size. From the perspective of epoch analysis, as the number of epochs training on each selected client at each communication round increases, the system can achieve significantly better test accuracy (Fig.~\ref{fig:HCFL-epoch-accuracy}) and loss (Fig.~\ref{fig:HCFL-epoch-loss}) after each round. To be more specific, the system in which the number of epochs is set at $20$ can converge after few rounds of communications. Meanwhile, the system with a minimum setting of training epoch gets saturated at the accuracy of around $86\%$ and the loss of $0.4$. The number of epochs affects the number of iterations that the \textcolor{duongfix}{model completes the training throughout} the entire training dataset. \textcolor{duongfix}{Hence, the predictor can achieve the better performance when we train the model with more epochs.} However, when the number of epochs becomes too large, the system can suffer from overfitting and extreme computational burden. \textcolor{duongfix}{In this work, we use five epochs for a balanced trade-off between the system computation and the system performance since the convergence of that setting is close to the capability of the pre-mentioned high epoch setting.}

In the following analysis, the batch size is taken into account. As observed from Fig.~\ref{fig:HCFL-batch-accuracy} and Fig.~\ref{fig:HCFL-batch-loss}, HCFL-integrated FL can achieve better performance when the small batch size is applied. The test loss reflects the categorical cross-entropy between the original and predicted labels of the data collected on clients. The maximum batch size can hardly attain the expected efficiency with a low-par accuracy and loss after $300$ communication rounds ($60\%$ and more than one respectively). It can be observed that, the setting of batch size can make a big contribution to the computation load of every client. Reducing the batch size helps the networks train faster and requires less memory with mini-batches. Moreover, the weight updates have more variance when applying a smaller batch size than when we use the larger batch size. This noise can act as a regularizing effect in weight updating, making the FL operate with better performance.

\textcolor{fix}{We make all our code publicly available for reproduction of results and extension \footnote[1]{\url{https://github.com/skydvn/hcfl-compression}}.}

\section{Conclusion and Future Works} 
\label{sec:conclusion}
A new compression scheme for FL was developed in this study. In particular, the proposed technique possesses several advantages over the conventional technique, such as high compression proficiency, high accuracy, and viability for the FL process (especially the FedAvg algorithm, wherein it is embedded at the server). For example, when the HCFL-integrated FL was set at the high compression ratio (i.e., $1$:$32$), the global test accuracy of the FL process was reduced by approximately $1\% - 2\%$ in comparison to the baseline method. It was confirmed that the HCFL can be implemented to compensate for the susceptibility to the error of the autoencoder reconstruction output. Further, the low complexity of the proposed compression scheme makes it highly suitable for the FL process, especially the state-of-the-art machine type communication systems such as unmanned aerial vehicle and smart city applications.  

In future work, we aim to improve the HCFL proficiency and reduce the supervised-training dependencies by applying deterministic algorithms on the encoder phase. Thus, HCFL is then only needed to train the decoder to adapt to the encoder's behavior. Moreover, multi-task learning is a promising method to aid the autoencoder's performance when implemented on deeper deep models such as ResNet, DenseNet, AlexNet, or VGG-16. Hence, HCFL can further be applied to a wide range of FL concepts.

\appendices 
\textcolor{fix}{
\section{Proof on Theorem~\ref{theorem:FL-distortion-rate}}
\label{appendix:proof-on-theorem-FL-distortion-rate}
Taken account of probability, the neural network is demonstrated in a synthesized model with many layers of hidden causal variables \cite{10.5555/2976456.2976476}. 
\begin{equation}\label{DBN0}
\begin{split}
P(x,g^1, \dots, g^l) &= P(x|g^1)\dots P(g^{l-2}|g^{l-1})P(g^{l-1},g^l),
\end{split}
\end{equation}
where all the conditional layers \(P(g^l|g^{l+1})\) are factorized distributions which are easy for computation of probability and sampling. Alternatively, each of the hidden layer can be represented by: 
\begin{equation}\label{DBN1}
\begin{split}
P(g^l|g^{l+1}) &= \prod^n_{j=1}{P(g^l_j|g^{l+1})}. 
\end{split}
\end{equation}
The model of the autoencoder consequently can easily be described through a Markov Chain with a close form of a system model similar to that of the Kalman filter, which can be built on linear operators that are perturbed by errors including Gaussian noise \cite{fundamentals-kalman-filtering}. It may be assumed that the multi-layer autoencoder model has a series of \(\hat{x}_i\) as output values in discrete time space,  where $i$ is the index of the discrete-time event. \(\hat{x}_i\) is the total of the original data input \(x_i\) with Gaussian noise \(v_i\), which is IID and drawn from a zero-mean distribution as follows: 
\begin{equation}\label{AE1}
\widetilde{x}_i = x_i + v_i. 
\end{equation}
The aggregation model proposed in \eqref{PMFA2} was used to aggregate the decoded weights at the server according to the following equation:
\begin{align}\label{AE2}
\widetilde{w}_t 
& = \frac{1}{K}\sum^{K}_{k=1}{\widetilde{w}_t^k} = \frac{1}{K}\sum^{K}_{k=1}{(w^k_t + v^k_t)} \notag\\
& = \frac{1}{K}\sum^{K}_{k=1}{w^k_t} + \frac{1}{K}\sum^{K}_{k=1}{v^k_t}
  = w_t + \frac{1}{K}\sum^{K}_{k=1}{v^k_t}.
\end{align}
A set of noise data at weight $t$ is given as: $V_t = \{v^1_t, \dots, v^K_t\}$. Assume \(S_K = \sum^{K}_{k=1}{v^k_t}\) the following equation is obtained: 
\begin{equation}\label{LLN0}
\begin{gathered}
E\left(\frac{S_K}{K}\right) = \mu. 
\end{gathered}
\end{equation}
 \(\mu\) is the expected value of the distribution \(V\), and \(E(\cdot)\) denotes the expectation function in \eqref{LLN0}. Furthermore, $V_t$ is regarded as a sequence of serially uncorrelated random variables because of their white noise property. Therefore, we have $\text{Cov}(v_t^i,v_t^j)=0$ ($\forall i,j$), where $\text{Cov}(v_t^i,v_t^j)$ is the covariance of two noise values corresponding to user $i,j$, and sampled from the set of noise data $V_t$. Hence, we have the following function:
 \begin{equation}\label{LLN1}
\begin{split}
E\left(\frac{S_K}{K} - \mu\right)^2 
    &= E\left(\frac{{S_K}^2}{K^2}\right) - {E\left(\frac{S_K}{K}\right)}^2     \\
    &= \frac{1}{K^2}\left[E\left({S_K}^2\right) - {E\left(S_K\right)}^2\right] \\
    &= \frac{1}{K^2}\left(\text{var}({S_K})\right) \\
    &= \frac{1}{K^2}\left[\sum^{K}_{i=1}{\text{var}(v^i_t)}+\sum^{K}_{i,j=1}{\text{Cov}(v_t^i,v_t^j)}\right]\\
    &= \frac{1}{K^2}\left(\text{var}(v^1_t) +\ldots + (\text{var}(v^K_t)\right),
\end{split}
\end{equation}
and the variance of each $v^k_t$ can be formulated as 
\begin{equation}\label{VAR0}
\begin{gathered} 
\text{var}(v^k_t) = \frac{1}{K}\sum^{K}_{k=1}{(v^k_t)^2} - \left(\frac{1}{K}\sum^{K}_{k=1}{(v^k_t)}\right)^2. 
\end{gathered}
\end{equation}
Applying the functions (\ref{LF1}) and (\ref{AE1}) into (\ref{VAR0}) yields: 
\begin{equation}\label{VAR1}
\begin{split} 
\text{var}(v^k_t) &= \frac{1}{K}\sum^{K}_{k=1}{(v^k_t)^2} - \left(\frac{1}{K}\sum^{K}_{k=1}{(v^k_t)}\right)^2 \\
       &\leq \frac{1}{K}\sum^{K}_{k=1}{(v^k_t)^2} 
       = \frac{2}{K}\mathcal{L}(w).
\end{split}
\end{equation}
From (\ref{LLN1}) and (\ref{VAR1}), we have 
\begin{equation}\label{LLN3}
\begin{split}
E\left(\frac{S_K}{K} - \mu\right)^2 
    &\leq \frac{2}{K^2}\mathcal{L}(w).
\end{split}
\end{equation}
It can be observed that, the function (\ref{LLN3}) gets converged at zero when the number of FL users \(K\) approaches infinity. It is evident that $\frac{S_K}{K}$  also converges at $\mu$. As \(v_i\) is drawn from a standard normal distribution, \(S_K\) is also drawn from the same distribution and has the mean value \(\mu = 0\). We have: 
\begin{equation}\label{VAR2}
\begin{gathered} 
\lim_{K \to \infty}{\frac{S_n}{K}} = \lim_{K \to \infty}{\frac{1}{K}\sum^{K}_{k=1}{v^k_t}} =0.
\end{gathered}
\end{equation}
Thus, when the number of clients in the FL model is large enough, the parameters aggregated at the server (\ref{PMFA2}) converge at their expected values:
\begin{equation}\label{VAR3}
\begin{split} 
\lim_{K \to \infty}{w_t} &= \lim_{K \to \infty}{\frac{1}{K}\sum^{K}_{k=1}{\widetilde{w}^k_t}} \\
&=\lim_{K \to \infty}{\left(\frac{1}{K}\sum^{K}_{k=1}{w^k_t} + \frac{1}{K}\sum^{K}_{k=1}{v^k_t}\right)} \\ 
&=\lim_{K \to \infty}{ \frac{1}{K}\sum^{K}_{k=1}{w^k_t}}.
\end{split}
\end{equation}
The following paragraphs discuss the relationship between the number of clients, the loss function of autoencoder and the accuracy rate of the aggregated data at the server. Applying Chebyshev’s inequality [17] into (13), we have the following formulation to prove the aggregated model convergence:
\begin{equation}\label{CERTAINTY0}
\begin{split} 
P\Big(\abs{\widetilde{w}_t -w_t} < \alpha\Big) &= P\left(\abs{\frac{S_K}{K} - \mu} < \alpha \right),
\end{split}
\end{equation}
where \(\alpha\) denotes the desired low-threshold of the weight deviation on each client. We can take the aggregated model accuracy as the certainty when $\abs{\widetilde{w}_t -w_t}$ is larger than a determined value $\alpha$ in the left side of the function. Applying the inequality in \cite{markov_process} yields
\begin{equation}\label{CERTAINTY1}
\begin{gathered} 
P\Big(\abs{\widetilde{w}_t -w_t} \geq \alpha\Big) \leq \frac{1}{\alpha^2}E\left(\frac{S_K}{K} - \mu\right)^2.
\end{gathered}
\end{equation}
Apply the inequality (\ref{VAR1}), we have an upper boundary for the certainty when the deviation of each  parameter is larger than a predefined threshold:
\begin{equation}
\begin{gathered}
P\Big(\abs{\widetilde{w}_t -w_t} \geq \alpha\Big) \leq \frac{2}{(K\alpha)^{2}}\mathcal{L}(w).
\end{gathered}
\end{equation}
\section{Proof on Theorem~\ref{theorem:FL-model-complexity}}
\label{appendix:proof-on-theorem-model-complexity}
Denote $H(\cdot)$ to be the entropy of the given dataset and $I(\cdot;\cdot)$ as the mutual information between 2 given sets of data, we have 
\begin{align}\label{AppendixA-1}
H(C)
& = I(W;\hat{W}) = H(W)-H(W|\hat{W}) \notag\\
& = H(W)-H(W-\hat{W}|\hat{W}) \notag\\ 
& \approx H(W) - H(W-\hat{W}), 
\end{align}
where $W$ and $\hat{W}$ are the sets of original model parameters $W = \{w_1, \dots, w_N\}$ and reconstructed model parameters $\hat{W} = \{\hat{w_1}, \dots, \hat{w_N}\}$, respectively. $C$ = $\{c_1, \dots, c_M\}$ is the set of compressed data with $N$ and $M$ being the sizes of the model parameters and compressed data, respectively. 
\newline
Because the HCFL consists of two sequential neural networks, the proposed model can be demonstrated in a Deep Belief Nets \cite{10.5555/2976456.2976476} structure where each neural node in the network contributes an independent probability to the output of the network's distribution:
\begin{align}\label{AppendixA-2}
P(\hat{W},W) 
& = \prod^{l}_{i=1}{P\left(g^i|g^{i+1}\right)} 
= \prod^{l}_{i=1}{\prod^{n_i}_{j=1}{P\left(g^i_j|g^{i+1}\right)}}, 
\end{align}
where $i$ and $j$ are the layer index of the neural network and the neural node index on each layer, respectively. The $l$ and $n_l$ denote the total layers and the number of nodes on each layer of the HCFL model, respectively. From \eqref{AppendixA-2}, we can assume that the noise generated during the HCFL process follows the Gaussian distribution $\mathcal{N}\left(0,E{\left[(W-\hat{W})^2\right]}\right)$ (as described in \ref{appendix:proof-on-theorem-FL-distortion-rate}). The $E{\left[(W-\hat{W})^2\right]}$ is the variance of the reconstructed parameter set $\hat{W}$. Therefore, the entropy of the loss between $W$ and $\hat{W}$ yields:
\begin{equation}\label{AppendixA-3}
\begin{split}
&~~~~H(W-\hat{W})= H\left(\mathcal{N}{\left(0,E{\left[(W-\hat{W})^2\right]}\right)}\right) \\
&= \frac{1}{2}\log{(2\pi{e})}E\left[{\left(W-\hat{W}\right)}^2\right] = N\log{(2\pi{e})}\mathcal{L}(w),
\end{split}
\end{equation}
where $H(\cdot)$ denotes the entropy of the equivalent information. Hence, we can estimate the loss function $\mathcal{L}$ in terms of the relationship between the input data and compressed data. We have
\begin{equation}
\begin{split}
\mathcal{L}(w) &\approx \frac{H(W)-H(C)}{N\log{(2\pi{e})}}\\
&= \frac{\sum_{N}P(w)\log{P(w)}-\sum_M{P(c)\log{P(c)}}}{N\log{(2\pi{e})}}.
\end{split}
\end{equation}
}
\bibliographystyle{IEEEtran}
\bibliography{Bib1.bib}
 
\end{document}